\RequirePackage{fix-cm}
\documentclass[smallextended]{svjour3}       
\smartqed  
\usepackage{graphicx}
\usepackage{amsmath}
\usepackage{mathptmx}      
\usepackage{enumerate}
\usepackage[misc]{ifsym}
\usepackage{graphicx}
\usepackage{fixltx2e}
\usepackage{lineno,hyperref}
\usepackage{graphicx}
\usepackage{epsfig}
\usepackage{subcaption}

\usepackage{caption}
\DeclareCaptionLabelSeparator{none}{ }
\usepackage[labelfont=bf,textfont=md]{caption}

\usepackage{pdfpages}
\usepackage{booktabs}
\usepackage{enumerate}
\graphicspath{ {./images/} }
\journalname{Mathematical Geoscience}

\usepackage[authoryear]{natbib}
\bibpunct{(}{)}{}{a}{}{;}
\setcitestyle{citesep={;}}

\newcommand{\tindog}{Region~A}
\newcommand{\homestead}{Region~B}
\newcommand{\tindogandhomestead}{Regions~A and B}

\begin{document}

\title{A Machine Learning Approach for Material Type Logging and Chemical Assaying from Autonomous Measure-While-Drilling (MWD) Data}

\author{Rami N.~Khushaba
	\and Arman Melkumyan
	\and Andrew J. Hill
}


\institute{All authors are within the Rio Tinto Centre for Mine Automation, Australian Centre for Field Robotics,
	The University of Sydney \at
	8 Little Queen street, Chippendale, NSW 2008. \\
	Tel.: +61 2 9351 4209\\
	\email{Rami.Khushaba@sydney.edu.au, \\
		Arman.Melkumyan@sydney.edu.au, \\
		Andrew.Hill@sydney.edu.au} \\
		Corresponding author ORCID: https://orcid.org/0000-0001-8528-8979
}

\date{Received: Dec 2020 / Accepted: date}

\maketitle

\begin{abstract}
Understanding the structure and mineralogical composition of a region is an essential step in mining, both during exploration (before mining) and in the mining process. During exploration, sparse but high-quality data is gathered to assess the overall orebody. During the mining process, boundary positions and material properties are refined as the mine progresses. This refinement is facilitated through drilling, material logging, and chemical assaying. Material type logging suffers from a high degree of variability due to factors such as the diversity in mineralization and geology, the subjective nature of human-measurement even by experts, and human errors in manually recording results. While laboratory-based chemical assaying is much more precise, it is time-consuming and costly and does not always capture or correlate boundary positions between all material types. This leads to significant challenges and financial implications for the industry, as the accuracy of production blast-hole logging and assaying processes is essential for resource evaluation, planning and execution of mine plans.

To overcome these challenges, this work reports on a pilot study to automate the process of material logging and chemical assaying. A machine learning approach has been trained on features extracted from measure-while-drilling (MWD) data, logged from autonomous drilling systems (ADS). MWD data facilitates building profiles of physical drilling parameters as a function of hole depth. A hypothesis is formed to link these drilling parameters to the underlying mineral composition. The pilot study results in this paper demonstrate the feasibility of this process, with correlation coefficients of up to 0.92 for chemical assays and 93\% accuracy for materials detection, depending on the material or assay type and their generalization across the different spatial regions. The achieved results are significant, showing opportunities to guide further drilling processes, provide chemistry data with a down-hole resolution, and continuously update mine plans as the mine progresses.
\end{abstract}


\section{Introduction}
\label{intro}

The mining industry has embarked on a journey to use automation to improve the accuracy and consistency of mining processes within its surface mines. Blast-hole drilling is a key task in most surface mining, as the accuracy of the information obtained affects the entire downstream mining process. This has impacts on i) scheduling, ii) excavation, iii) slope stability, iv) material handling, v) beneficiation, vi) ore loss and vii) final product blending \citep{McHugh2012}. Measurement-while-drilling (MWD) enables the collection of accurate, fast and high-resolution information from production blast-hole drills for equipment automation and monitoring the health of major drilling items \citep{MWD2015}. In open-pit mining, MWD systems monitor several performance factors including, but not limited to, i) rate of penetration, ii) torque, iii) rotation pressure, iv) speciﬁc energy of drilling, v) weight on bit and vi) rotary speed. These factors are becoming standard features on the blast-hole drill rigs supplied by most equipment manufacturers \citep{Raymond001}. MWD data has already been utilized in applications related to i) rock type recognition \citep{FlorianRig1}, ii) boundary identification and surface updates \citep{Silversides}, iii) automated coal seam detection \citep{LEUNG2015196}, iv) estimation of rock mass rating during tunneling \citep{MWDRock}, v) rock fracture density characterization \citep{Khorzoughi18}, vi) improving rock-breakage efficiencies \citep{PARK2020179}, vii) developing a prediction model of over- and under-excavation depths from blasting \citep{MWDBlasting}, viii) characterizing a coal mine roof \citep{KhanalMWD}, and ix) identifying the top of coal seams allowing drilling to be halted to prevent unintended blasting of coal and associated problems \citep{Raymond001}. However, an extensive literature review has not uncovered any use of MWD data to estimate material types or chemistry assays.

Boreholes and blast-holes are routinely logged and assayed to assist in the understanding of the structures and mineralogical compositions of an area \citep{Sommerville}, the accuracy of which is essential for resource evaluation and planning in the minerals industry. Material logging is the process of recording manual geological observations to identify the material types present in the sample, and chemistry assaying is generally a lab-based process to quantify the chemical make-up of samples (valuable minerals, impurities and water content).

To standardize primary characteristics, such as mineralogy and texture, in a hierarchical manner, the Material Type Classification Scheme was developed \citep{Box2002,WEDGE2019103118}. This scheme models physical and chemical attributes for predicting metallurgical behavior and the quality of the product for optimal ore processing \citep{Paine}. Logging of material-types is usually performed on chip samples produced by reverse circulation (RC) drills during exploration (in 2m intervals) or blast-hole cone samples (one sample per blast hole). This generates a huge number of samples and significant costs and time in preparation and assaying. For each sample, a geologist manually handles the material and estimates several parameters including \citep{Sommerville,DanielAVA}: (i) the percentages of various material types present in the sieve (usually in increments of 5\%), (ii) the sample color, (iii) the shape of the chips, and (iv) the percentage of material recovered. Another sample is usually sent for laboratory X-ray fluorescence assay analysis to measure i) aluminium oxide (Al\textsubscript{2}O\textsubscript{3}), ii) iron (Fe), iii) silicon dioxide (SiO\textsubscript{2}), iv) phosphorus (P), v) sulfur (S), vi) manganese (Mn), vii) magnesium oxide (MgO), viii) titanium dioxide (TiO\textsubscript{2}), ix) calcium oxide (CaO), and x) total loss on ignition (LOI). However, this approach has limitations as exploration holes are sparse in the horizontal direction, while horizontally denser blast-holes provide only a single sample for the entire hole (e.g.~10-15m depth). For these reasons, the literature criticizes assaying from RC boreholes or blast-hole samples in that they do not always provide truly representative analysis \citep{Stats}, are not always effective at characterizing thin layers, and the sampling is costly to perform. As a result, there has been an increased interest in downhole, in-situ assaying. Downhole Pulsed Fast and Thermal Neutron Activation (PFTNA) has been proven through its operations to date to provide a downhole in-situ assaying technique that has several advantages over conventional sample-based assays in terms of safety, cost, cycle-time of results, and accuracy \citep{IRONORE20192,IRONORE2019}. By removing the need for manual samples taken on-site, tools relying on PFTNA technology limit the exposure of workers to manual handling, operational and environmental safety risks typical on many mine sites.

It is important to mention here that the use of PFTNA-based technologies alone does not completely obviate the need for manual handling for logging purposes, as such technologies are primarily used for assaying, while material logging is not facilitated by PFTNA. It has been reported that using PFTNA for in-situ measurements, as an alternative to traditional chemical analysis, may be overly-simplistic, as it does not provide some vital information such as hardness \citep{PFTNA001}. Additionally, PFTNA does not produce estimation for all assays, especially trace elements which can be of particular importance in some deposits, such as P in iron ore, or Au and Ag in copper.

This paper argues that improvements in drilling technology can change the way mineral deposits are observed. The motivation of this work is driven by the fact that MWD data has already been deemed useful for several applications \citep{FlorianRig1,Silversides,LEUNG2015196,MWDRock,Khorzoughi18,PARK2020179,MWDBlasting,KhanalMWD,Raymond001}, which lends itself as a potential tool for determining the material-types and chemical assays. Hence, the primary hypothesis of this work is that using MWD and machine learning can help predict chemical assays and materials. This is a novel application of MWD data, which has not been previously verified and is based on a common conclusion that the drilling behavior is related to the mechanical properties of the material being drilled. The result would be a downhole in-situ assaying or logging ``equipment-as-a-sensor'' tool that lends itself particularly well to bulk mining operations such as that found in the iron ore operations in the Pilbara \footnote{https://en.wikipedia.org/wiki/Pilbara}. Additionally, MWD data has good down-hole resolution and can be made available in real-time, during drilling. A key challenge is the difficulty modelling the bit–rock interaction because of high-order variability in different rocks or even simply at different sample points in the same rock material due to the presence of cracks, ﬁssures and a host of other discontinuities \citep{MWD2015}. This is where the use of machine learning (ML) models to capture the relation between the MWD data and materials or assays become increasingly important. The application of ML in this problem is supported by wide-ranging successes of using ML in several applications in the era of big data \citep{MLBigData}.

\section{Data Sources}

This pilot study was conducted on an Iron Ore mine site in the Hammersley Range, which is located in the Pilbara region of Western Australia. MWD, material logging and chemistry assay data was collected from two physically separate regions, labelled here as \tindogandhomestead{}.

\subsection{MWD Data}
\begin{figure}[ht]
\centering
\includegraphics[width=8cm]{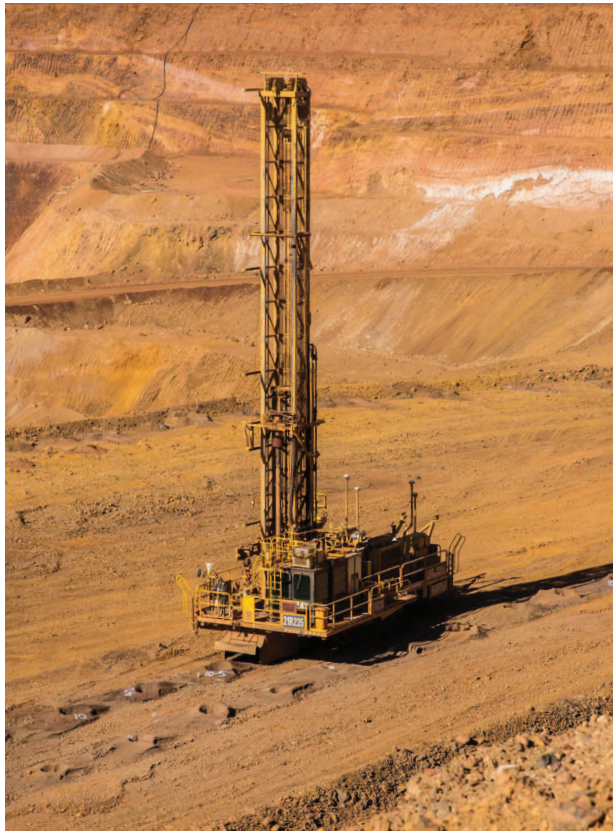}
\caption{Autonomous blast-hole drill rig used for collecting experimental MWD data}
\label{Fig001}
\end{figure}

MWD data represents real-time measurements of several mechanical signals, collected from sensors equipped on relatively large drill-rigs used in mining for blast-hole drilling. These signals are commonly used to control and monitor the performance of drilling. \figurename{ \ref{Fig001}} shows the autonomous blast-hole drill rig that collects the MWD data used in this paper. The parameters of interest considered in this paper include: i) the time taken for hole development (start to finish), ii) measurement depth, iii) average head rotation speed (rotationRPM), iv) average bit air pressure (airPressure), v) average feed pressure (feedPressure), vi)  average torque (torque), vii) average rate of penetration (rop), viii) average force on bit (fob), ix) average rotation pressure (rotationPressure), x) adjusted penetration rate (apr), and xi) specific energy of drilling (sed). Also considered is the ratio of rotationPressure to feedPressure as another parameter. The MWD time-series data were logged by the drilling system as discretized sequences, typically discretised into depth segments of $0.1$ meter resolution \citep{Raymond001}. A total of almost 7000 holes' data were logged by the ADS system, as MWD is available for every hole. This is not the case for materials/assays data as mentioned in the next section.

\subsection{Logging and Assaying}
\begin{figure*}[ht]
	\centering
	\includegraphics[width=12cm]{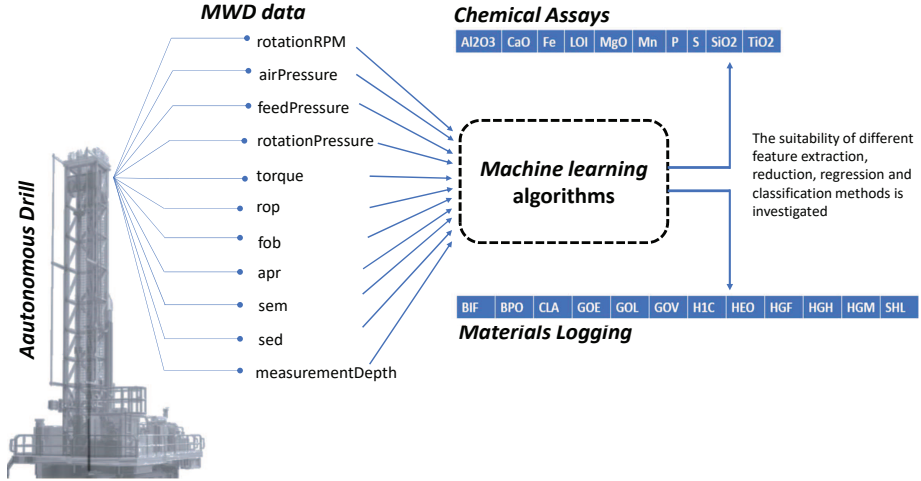}
	\caption{The three sources of data available for this pilot study. MWD is sampled at 0.1m intervals across the entire hole depth, while assay and material types offer a single value per hole for each parameter, for the subset of holes where they are collected}
	\label{Fig002}
\end{figure*}

The logged material-types and chemistry assay data were provided from multiple blasts within both Region A and Region B. While the autonomous drilling system provided MWD data along the full depth of every blast-hole, the data for material-types and chemistry is only available for a subset of the holes, and where available is a single value per hole for each material and chemistry parameter (depicted in \figurename{\ref{Fig002}}). This is primarily due to the cost of data collection, including manual logging, manual physical sampling and laboratory analysis costs. Additionally, this data is delayed by the collection and analysis time, while MWD data is collected electronically and is immediately available after drilling.

The resulting data set comprises MWD for all holes, and subsets of holes that also include material logging and/or chemistry assays, as noted in Table~\ref{tab:datacounts}. 

\begin{table}
	\centering
	\begin{tabular}{lccc}
		Data Sources & \tindog{} & \homestead{} & Total \\
		\hline
		MWD + Material & 805 & 993 & 1,798 \\
		MWD + Chemistry & 2,176 & 2,408 & 4,584 \\
		MWD + Mat.~+ Chem. & 537 & 804 & 1,341 \\
		\hline
	\end{tabular}
	\caption{Material logging and chemistry assays are collected more sparsely than MWD data, resulting in multiple subsets of holes with different combinations of data for the same hole}
	\label{tab:datacounts}
\end{table}

Additionally, wherever material logging is available, a theoretical chemistry assay is also estimated by geologists. A validation process is usually performed by geologists after logging, adjusting the percentages of the logged material types or adding/removing material types where necessary, using geologically informed substitutions so that the theoretical assay values are within an error tolerance of the laboratory assay values obtained from the interval’s chip samples \citep{DanielAVA}.

\section{Details of the Machine Learning Pipeline}

\subsection{Feature Extraction}
The literature review showed that there is no generally agreed-upon quantitative method for the description of the MWD data. As such, several feature extraction methods are investigated in detail in this study. As our hypothesis attempts to relate the changes in the drilling parameters to the changes in materials and chemical assays underground, we have sought features focusing on changes in the amplitude, energy, complexity, frequency contents and shape of the power spectrum, signals' peak values, root mean square and entropy of the signals to denote the dispersion. The selected methods have been utilized across different time-series data in other domains, demonstrating significant success in describing signals \citep{HJORTH1970306,KHUSHABA201442,ToonSSI}, and hence are seen as good candidates to employ to describe the MWD signals. The investigated feature extraction methods are described below. In the following features' descriptions, we assume a signal $\textbf{x}$, with $x[j]$ representing its $j$-th sample, where $j = 0,2,3,...,N-1$, with a length of $N$ samples.

\textbf{Hjorth parameters}: 

These parameters were originally developed in an approach to describe the shape of the frequency spectrum of any signal directly from the time-domain by using Parseval's theorem and Fourier transform properties \citep{HJORTH1970306}. Several spectral analysis methods (fast Fourier transform, wavelets transform, Wigner-Ville transform, plus many others) can be used to study the frequency contents of the underlying signals. However, Hjorth parameters were chosen because of their low computational cost which makes them favorable for any online data processing task \citep{KHUSHABA201442}. The description of these parameters starts by first observing Parseval's theorem, which states that the sum of the square of the function is equal to the sum of the square of its transform

\begin{equation}
\sum^{N-1}_{j=0} x[j]^2 = \frac{1}{N} \sum^{N-1}_{k=0} \left|X[k]X^*[k]\right| = \sum^{N-1}_{k=0} P[k],
\end{equation}

\noindent where $X[k]$ is the discrete Fourier transform (DFT) of the original signal $\textbf{x}$, $P[k]$ is the phase-excluded power spectrum, that is the result of a multiplication of $X[k]$ by its conjugate $X^{*}[k]$ divided by $N$, and $k$ is the frequency index. As the complete frequency description derived by the Fourier transform is symmetric with respect to zero frequency (having identical branches stretching into both positive and negative frequencies) \citep{HJORTH1970306}, all odd moments will become zero. This is according to the definition of a moment $m$ of order $n$ of the power spectral density of the signal $\textbf{x}$ which is given by

\begin{equation}
m_n = \sum^{N-1}_{k=0} k^n P[k].
\label{eq001}
\end{equation} 

Hjorth parameters are hence mainly based on the lower order even moments, denoted as $m_0$, $m_2$, and $m_4$:

\begin{equation}
m_0 = \sum^{N-1}_{k=0} k^0 P[k] = \sum^{N-1}_{j=0} x[j]^2, 
\end{equation} 

\begin{equation}
m_2 = \sum^{N-1}_{k=0} k^2 P[k] = \frac{1}{N} \sum^{N-1}_{k=0} (k X[k])^2 = \frac{1}{N} \sum^{N-1}_{j=0} (\Delta x[j])^2,
\end{equation}

\begin{equation}
m_4 = \sum^{N-1}_{k=0} k^4 P[k] = \frac{1}{N} \sum^{N-1}_{k=0} (k^2 X[k])^2 = \frac{1}{N} \sum^{N-1}_{j=0} (\Delta^2 x[j])^2,
\end{equation}

\noindent where $\Delta^n$ is the $n$-th derivative of a function in the time-domain, this is according to the time-differentiation property of the Fourier transform. This property states that the power spectrum of a signal in the frequency domain multiplied by $k$ raised to the $n$-th power is equivalent to $n$-th derivative of the same signal in the time-domain. Hjorth parameters of interest are then calculated based on the above moments as shown below

\begin{equation}
\textrm{Activity}  = m_0, 
\end{equation}
\begin{equation}
\textrm{Mobility}  = \sqrt{\frac{m_2}{m_0}}, 
\end{equation}
\begin{equation}
\textrm{Complexity}  = \frac{\sqrt{\frac{m_4}{m_2}}}{Mobility},  
\end{equation}

\textbf{Waveform Length (WL)}: 

This is an intuitive measure to describe the cumulative length of the waveform over the considered segment which can also indicate the waveform complexity. The resultant value of the $WL$ calculation also indicates a measure of the waveform amplitude, frequency, and duration \citep{Hudgins1993}. For a given hole depth, the WL feature value grows larger for signals with higher frequencies than those with lower ones. This is given by

\begin{equation}
WL = \sum^{N-2}_{j=0} |x[j+1] - x[j]|,
\end{equation}

\textbf{ Simple Square Integral (SSI)}: 

SSI captures the energy of the signal under consideration as a feature \citep{ToonSSI}. It can be expressed as

\begin{equation}
SSI = \sum^{N-1}_{j=0} |x[j]|^2,
\end{equation}

\textbf{Crest Factor (CF)}: 

CF is defined as the ratio of the absolute value of the peak in the signal under consideration divided by the Root Mean Square (RMS) value of the same signal \citep{ShockBook}. The crest factor indicates how extreme the peaks are in a waveform, with CF of 1 indicating no peaks and larger values indicating more peaks. CF also expresses the size of the dynamic range for an input signal. This is given by

\begin{equation}
CF = \frac{|x_{peak}|}{x_{RMS}},
\end{equation}

\noindent where 

\begin{equation}
x_{RMS} = \sqrt{\frac{1}{N} \sum_{j=0}^{N-1} x[j]^2},
\end{equation}

\textbf{Pressure Ratio (PR)}: 

A new measure was created by taking the ratio of the rotationPressure to that of the feedPressure, this was inspired by the earlier work of \cite{LEUNG2015196}. This ratio forms a derived drill performance indicator, the characteristics of which this work attempts to link to the different materials type and chemical assays. These characteristics included all the following:

\begin{enumerate}[I]
\item summation of the absolute difference between consecutive samples of the PR 
\item summation of the PR squared (SPR2) and its logarithmically scaled version
\item summation of the first derivative of the PR squared (SDPR2)
\item summation of the second derivative of the PR squared (SDDPR2)
\item a logarithmically scaled version of SDPR2/SPR2
\item a logarithmically scaled version of SDDPR2/SDPR2
\item the maximum of PR multiplied by the max of $fob$.
\end{enumerate}

\textbf{Singular-Value Decomposition Entropy (SvdEn)}: 

The dispersion of the singular values $\lambda_k$ also indicates the complexity of the signal dynamics and is an indicator of how many vectors are needed for an adequate explanation of the signals being studied. To calculate SvdEn, the singular values are first normalized by their total summation

\begin{equation}
\bar{\lambda}_k = \frac{\lambda_k}{\sum \lambda_k}, 
\end{equation}

\noindent where $\sum \bar{\lambda}_k =1$. To denote the dispersion characteristics of the singular values, SvdEn is defined with the Shannon formula applied to the elements of singular values of the matrix and is calculated as shown below \citep{SvdEnt}.

\begin{equation}
\textrm{SvdEn} = - \sum  \bar{\lambda}_k \textrm{log} \bar{\lambda}_k,
\end{equation}

SvdEn can also be considered as a measure of feature-richness in the sense that the higher the entropy of the set of SVD weights, the more orthogonal vectors are required to adequately explain it.

\textbf{Signal Flatness (F)}: 

This is defined as the ratio between geometric and arithmetic means and is considered as an important measure to distinguish signals that are flat or do not change much, from those that have the amplitude concentrated across small ranges. A high signal flatness (approaching 1.0) indicates that the signal has a similar value across all samples, while a low flatness (approaching 0.0) indicates that the power is concentrated in a relatively small number of samples \citep{Johnston608,Shlomo}. 

\begin{equation}
\mathrm{Flatness} = \frac{\sqrt[N]{\prod_{j=0}^{N-1}x[j]}}{\frac{\sum_{j=0}^{N-1}x[j]}{N}} = \frac{\exp\left(\frac{1}{N}\sum_{j=0}^{N-1} \ln x[j]\right)}{\frac{1}{N} \sum_{j=0}^{N-1}x[j]},
\end{equation}

\textbf{Descriptive Statistics}: 

The following common statistical measures are also included as features: the signal maximum, standard deviation, skewness, kurtosis, mean, geometric mean, and median of each signal.

\subsection{Regression and Classification Models}
Several regression and classification models were utilized in this research. The Support Vector Machines (SVM), multivariate Gaussian Process (GP), and Random Forests (RF) models were all utilized for MWD versus assays regression analysis, while SVMs classification models were also utilized for the detection of the different material-types as described in the experiments section. The details of these models are omitted from this paper as these are traditional algorithms for which the description can be found in general pattern recognition references \citep{PRbook,GPbook}. These models were utilized with the extracted MWD features described in the previous section, with the output of the models being the estimation of the chemical assays for regressions models and the classification labels (material detected or not) for the material detection problem.

The testing scheme utilized with the above models employed a cross-validation process across two modes, these are: (i) random k-fold cross-validation \citep{crossVal001,crossVal002} and (ii) spatial k-fold cross-validation \citep{Roberts2017,MEYER2019108815,Talebi2020}. Overall, the goal of cross-validation is to test the model's ability to predict based on new data that was not used for training the models, to flag problems like overfitting or selection bias \citep{fivefold}, and to give an insight on how the model will generalize to an independent dataset (an unknown dataset). While the cross-validation procedure is the same for random and spatial cross-validation (the process of dividing the data into separate segments for training and testing), the major difference is how the data points are split into folds. The chosen mode of validation is hence largely dependent on the goal of the experiments.

In the case of random k-fold cross-validation, it is generally known that blast-hole sampling is a sparse process, where not all the drilled holes would be considered for logging physical materials or laboratory sampling of chemical assays. In this regard, the general approach is to sample every $n$-th hole, with $n$ varying across the mine sites depending on design specifications (roughly 1/6 or 1/8 for materials logging and 1/4 for assays sampling). A potential application of this work would be to fill these data gaps with inferenced materials and chemical assays, using the nearby holes where data has been recorded as the training set. This mimics the process of random k-fold cross-validation, and the short distance scales do not require spatial k-fold cross-validation for this demonstration. The random k-fold cross-validation experiments are included as a proxy for this use case.
	
In the case of spatial k-fold cross-validation, recent literature has shown that when using a model trained on spatial data to make inferences on data collected from a spatially distant environment, the commonly used random cross-validation may provide considerably over-optimistic error estimates due to the problem of spatial autocorrelation \citep{Roberts2017, MEYER20181}. Cross-validation strategies based on random data splitting fail to assess models' performance in terms of spatial mapping. If the objective is to test the model's performance upon spatially distant datasets, then a spatially-aware cross-validation scheme will provide more realistic outcomes to accurately estimate the generalization errors. In this regard, we have implemented a leave-one-blast-out scheme of testing in which the data from all holes belonging to a specific blast is kept away for testing, while the remaining data from all other blasts are used for training (a process that is repeated across all blasts). This mimics the use-case of predicting assays in a new blast in real-time from MWD data, before assays and material logs take place, or for selected blasts where these are not economically viable.
	

For most of the regression models, Bland-Altman graphs and QQ-plots are shown to verify the outcome of these models in comparison to the laboratory measurements. To verify the results statistically, the following measures are also shown on the correlation graphs: 

\begin{enumerate}[I]
\item \textbf{eq} - slope and intercept equation
\item \textbf{r} - Pearson r-value
\item \textbf{RMSE} - root mean squared error
\item \textbf{p} - Pearson correlation $p$-value
\item \textbf{n} - number of data points used
\item  \textbf{RPC} - reproducibility coefficient (1.96*SD)
\item \textbf{CV} - coefficient of variation (SD of mean values in \%)
\end{enumerate}

The Bland-Altman plot is presented as a scatter plot in which the x-axis represents the average of a pair of measurements $(A + B)/2$, and the y-axis shows the difference between the two paired measurements $(A - B)$. The Bland-Altman plot allows visual inspection for several aspects of the comparability of outcomes of the utilized model against the actual laboratory measurements. First, a consistent measure of bias can be described. This is the mean of all the differences between the algorithms' outcomes and the laboratory measurements. This mean is represented as a line across the x-axis of the plot, with the difference between this value and y-values describing the magnitude and direction of the bias. Bias can be reported in absolute terms or as a percentage (bias/mean value). The Bland-Altman plots are also utilized to study any proportional bias between the model outcomes and the laboratory measurements. The existence of proportional bias indicates that the model and laboratory measurements do not agree equally through the range of measurements (the limits of the agreement will depend on the actual measurement).

\section{Experimental Results}
In this section, the feasibility of predicting several assays from MWD data with the various regression models is investigated. The proposed approach is demonstrated on selected assay types, as a general proof of concept. The results in the following sections are acquired using the random k-fold cross-validation mode (as filling in the gaps was the primary motivator of this research), unless otherwise specified that the spatially-aware cross-validation mode was used.
 
\subsection{Iron (Fe) Prediction}
\begin{figure}[t!]
    \begin{subfigure}[t]{\columnwidth}
        \centering
						\includegraphics[width=10cm,trim={2cm 1cm .5cm 1cm},clip]{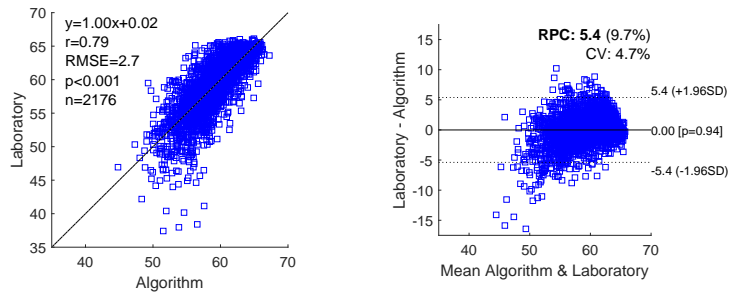} 
						\caption{GP model results}
    \end{subfigure}%
   
		\begin{subfigure}[t]{\columnwidth}
        \centering
						\includegraphics[width=10cm,trim={2cm 1cm .5cm 1cm},clip]{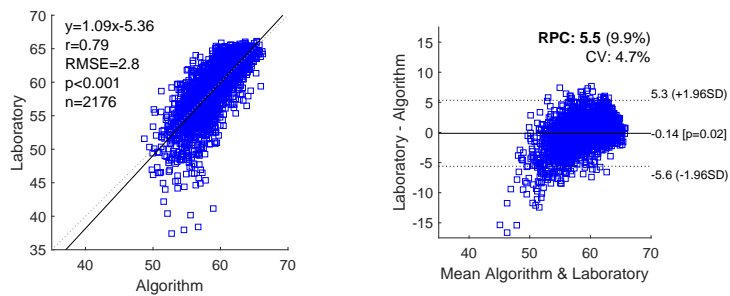} 
						\caption{SVM model results}
    \end{subfigure}%
	
    \begin{subfigure}[t]{\columnwidth}
        \centering
						\includegraphics[width=10cm,trim={2cm 1cm .5cm 1cm},clip]{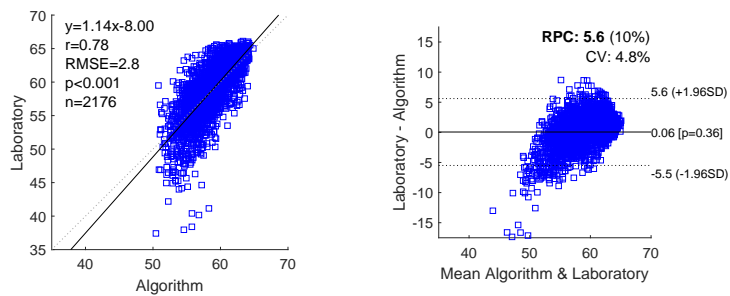} 
						\caption{RF model results}
    \end{subfigure}
    \caption{\tindog{} Bland-Altman iron (Fe) regression plots with several models}
		\label{Fig003}
\end{figure}

\begin{figure}[t!]
    \begin{subfigure}[t]{\textwidth}
        \centering
		    \includegraphics[width=10cm,trim={2cm 1cm .5cm 1cm},clip]{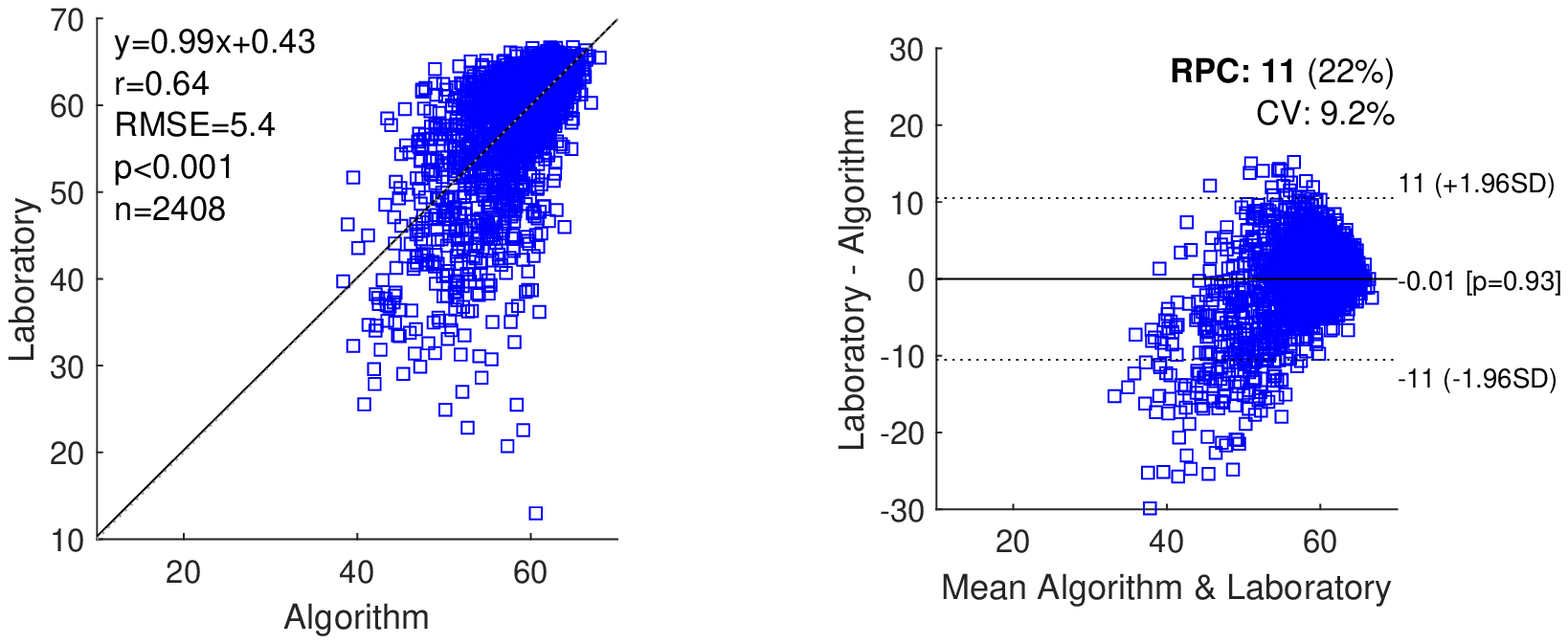} 
		    \caption{GP model results}
    \end{subfigure}%
		
		\begin{subfigure}[t]{\textwidth}
        \centering
        	\includegraphics[width=10cm,trim={2cm 1cm .5cm 1cm},clip]{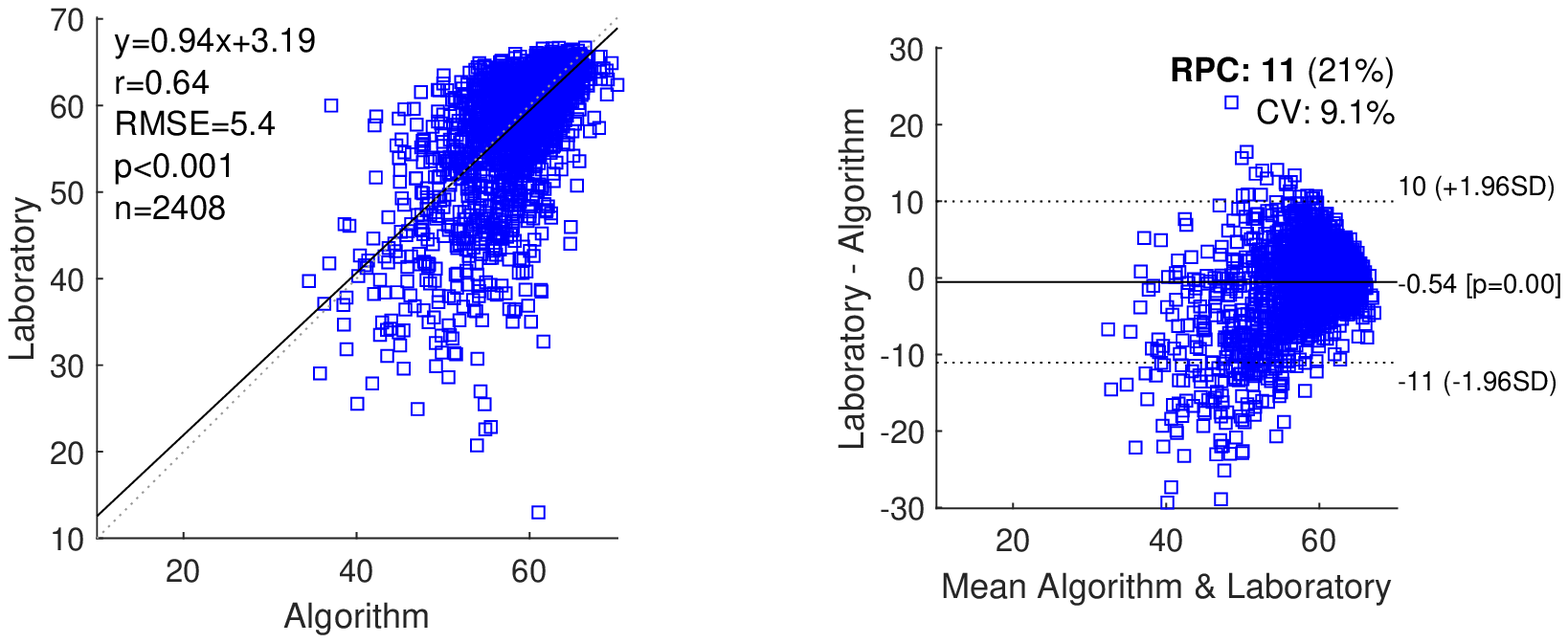} 
		    \caption{SVM model results}
    \end{subfigure}%
		
    \begin{subfigure}[t]{\textwidth}
        \centering
        	\includegraphics[width=10cm,trim={2cm 1cm .5cm 1cm},clip]{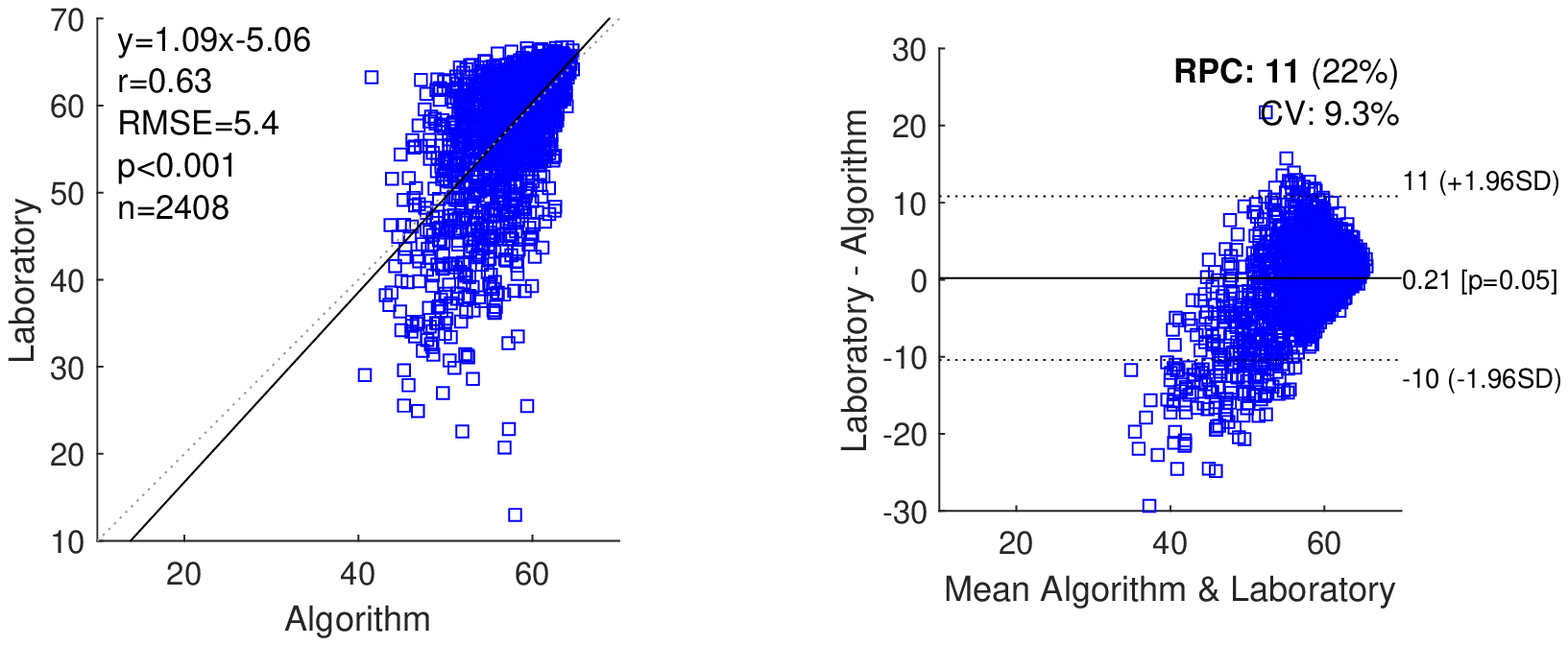} 
		    \caption{RF model results}
    \end{subfigure}
    \caption{\homestead{} Bland-Altman iron (Fe) regression plots with several models}
	\label{Fig004}
\end{figure}

The regression results for predicting iron (Fe) using GP, SVM, and RF models are shown in \figurename{ \ref{Fig003}} and \figurename{ \ref{Fig004}} for \tindogandhomestead{} respectively. For \tindog{}, both SVM and GP had a Pearson correlation coefficient of 0.79 with the laboratory readings for Fe, with an RMSE of 2.7 and 2.8 respectively for GP and SVM. RF performed similarly to GP and SVM, with a correlation coefficient of 0.78 and an RMSE value of 2.8. For \homestead{}, both GP and SVM again performed similarly, though with a lower Pearson correlation coefficient of 0.64, and RMSE of 5.4, while RF achieved a correlation coefficient of 0.63, and RMSE of 5.4. All regression models from both sites had $p$-values less than 0.001, which indicates a significant correlation between the Fe estimates from these models vs. that from the corresponding laboratory measurements. From these results, it is clear that all models achieved better results on \tindog{} data in comparison to the performance on \homestead{}. Therefore, further analysis to understand the differences between \tindogandhomestead{} were carried out.

\begin{figure*}[t!]
    \begin{subfigure}[t]{0.3\textwidth}
        \centering
		    \includegraphics[height=1.5in]{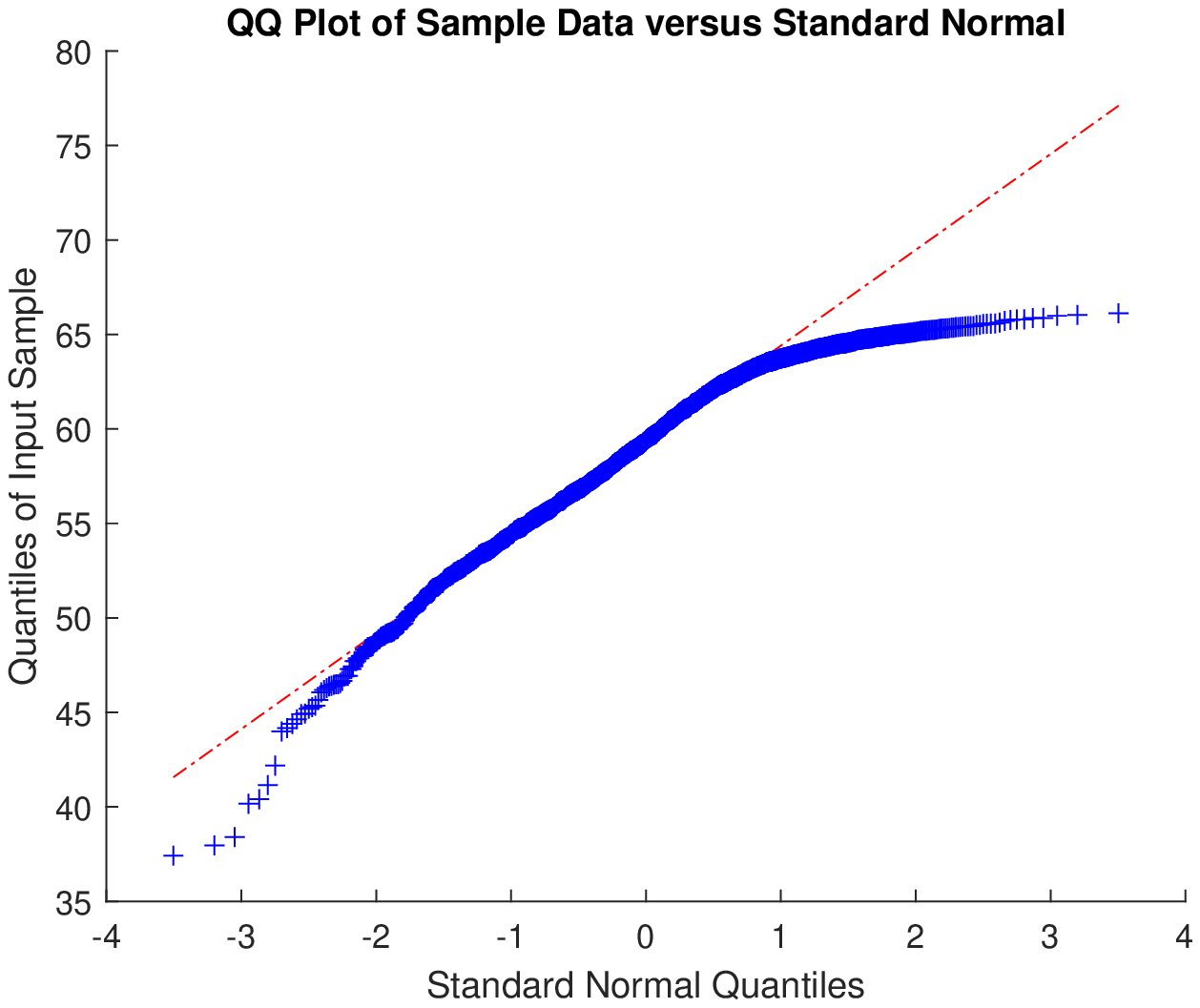} 
		    \caption{\tindog{} Fe}
    \end{subfigure}
   	~
	\begin{subfigure}[t]{0.3\textwidth}
        \centering
         	\includegraphics[height=1.5in]{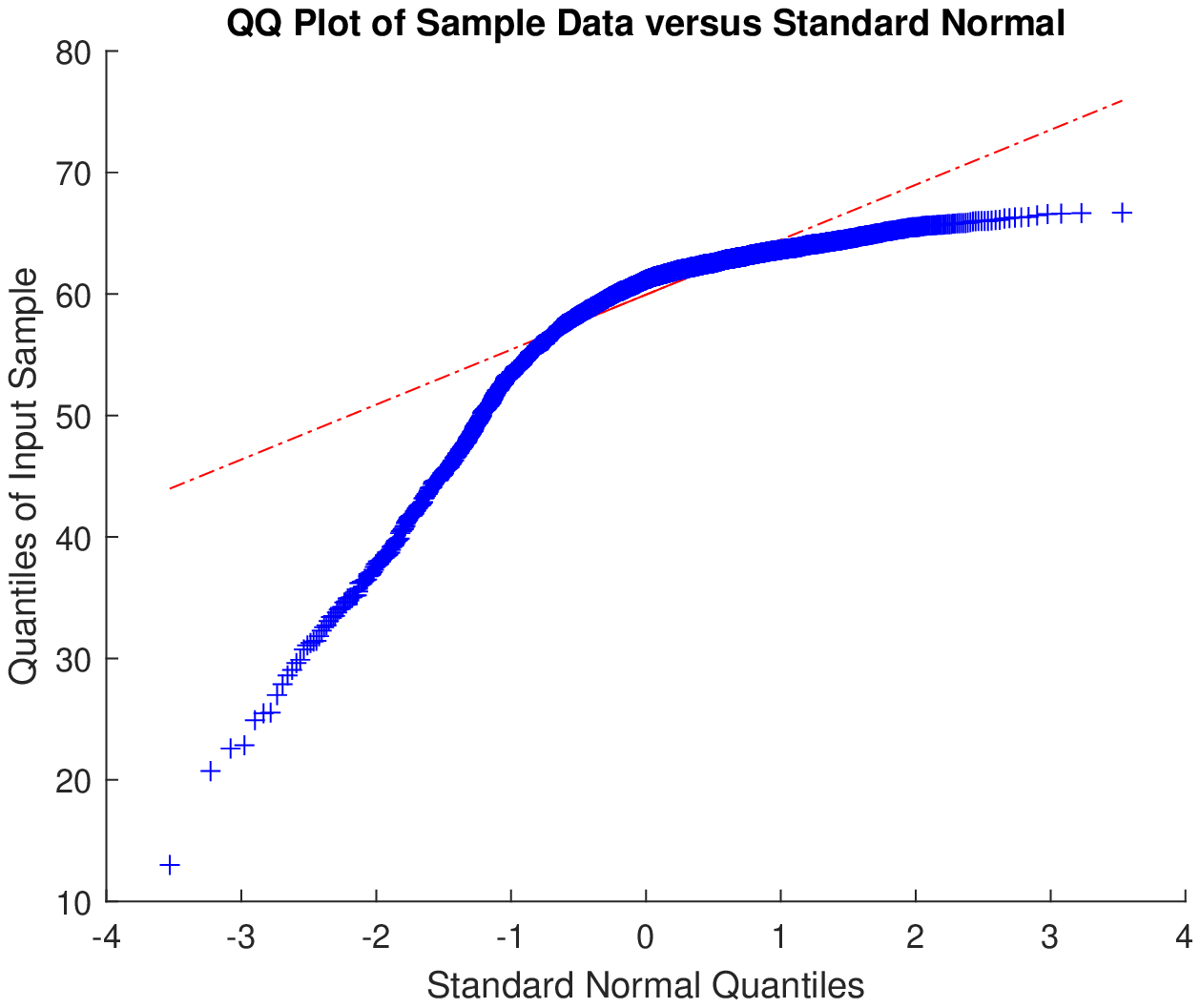} 
	     	\caption{\homestead{} Fe}
    \end{subfigure}
	~
	\begin{subfigure}[t]{0.3\textwidth}
        \centering
         	\includegraphics[height=1.5in]{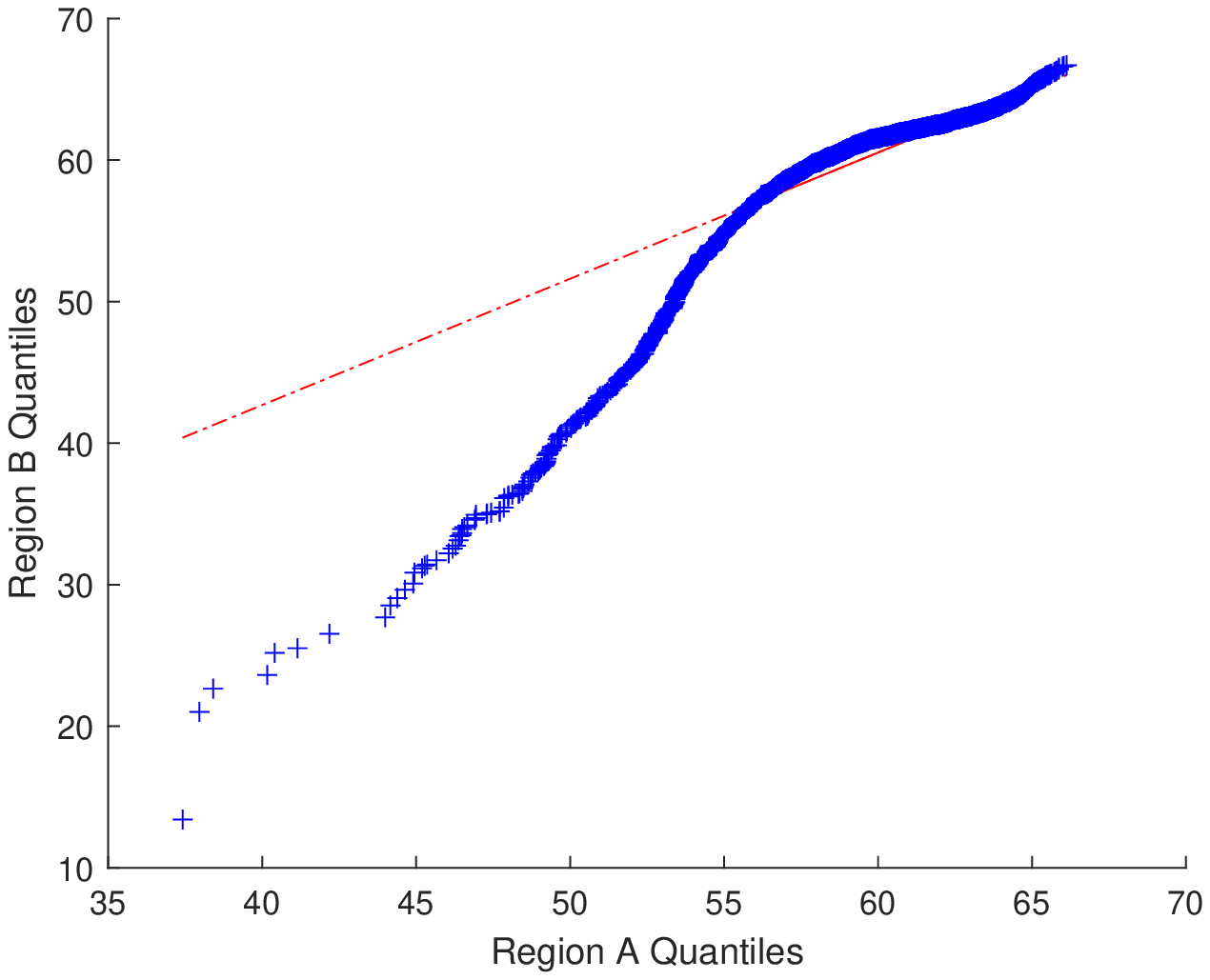} 
	     	\caption{\tindog{} Fe vs.~\homestead{} Fe}
    \end{subfigure}
    \caption{QQ-plots displaying empirical quantile-quantile plot for \tindogandhomestead{}. Samples from both mining sites are not normally distributed, and both samples do not come from the same distribution}
		\label{Fig005}
\end{figure*}

\begin{figure}[h]
	\centering
		\includegraphics[width=0.5\textwidth]{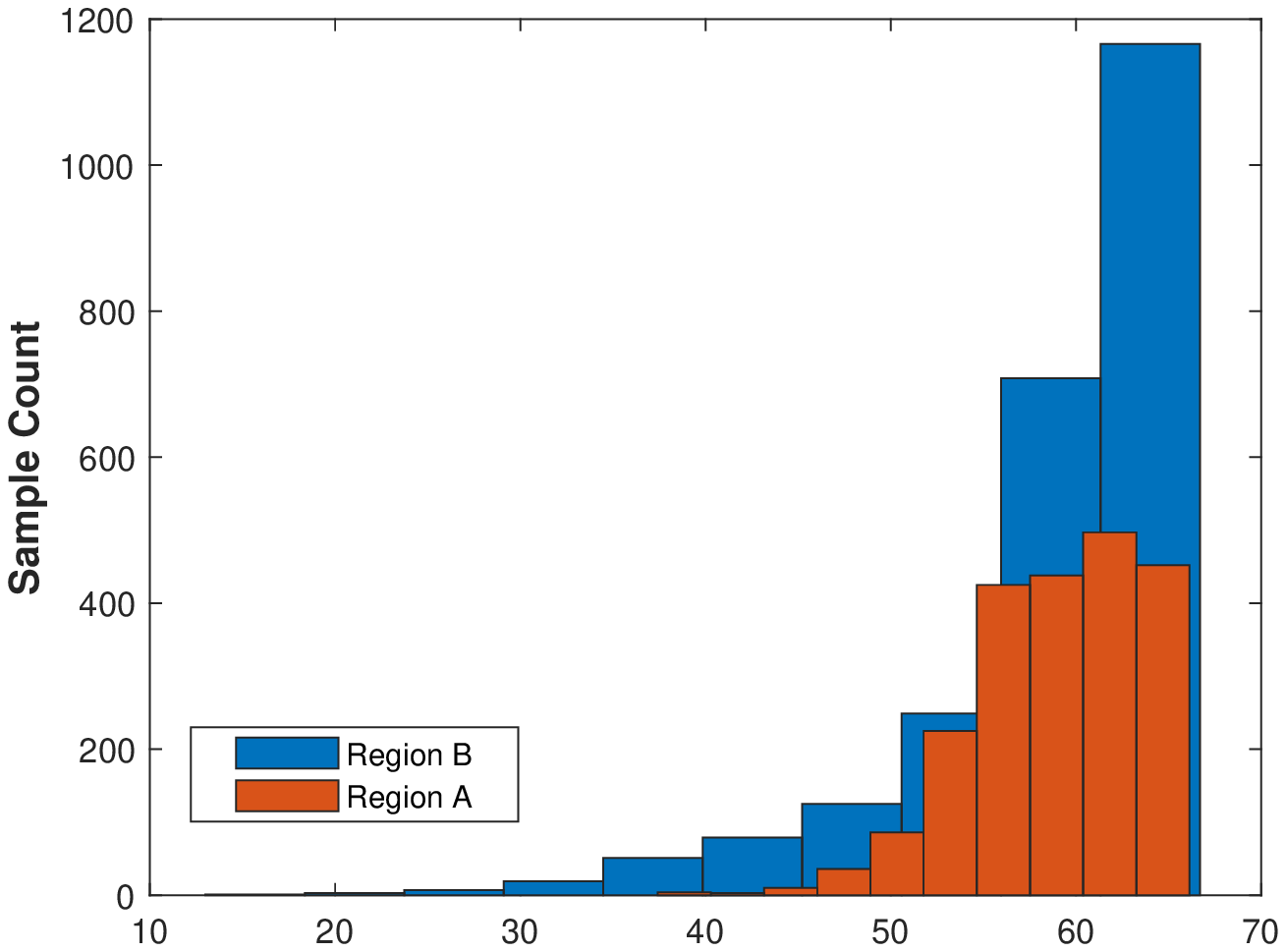}
	\caption{Histogram of Fe samples from \tindogandhomestead{} showing clear differences in distributions between the two mining sites}
	\label{Fig006}
\end{figure}

The distributions of the laboratory measurements of Fe from both sites were then studied to understand the differences between \tindogandhomestead{}. \figurename{ \ref{Fig005}} shows the QQ-plot for both sites individually against the quantiles of normal distribution, and also against each other's quantiles. The results in \figurename{ \ref{Fig005}} clearly demonstrate that Fe samples from each site are not normally distributed and that the samples of \tindog{} Fe do not come from the same distribution as \homestead{} Fe. Further evidence that these distributions are different is seen in the histogram of the Fe samples from each site, as shown in \figurename{ \ref{Fig006}}. Both histograms are skewed, but \homestead{} Fe exhibits twice the kurtosis of \tindog{} Fe.

\begin{figure*}[t!]
		\begin{subfigure}[t]{0.5\textwidth}
		\centering
		\includegraphics[width=6cm]{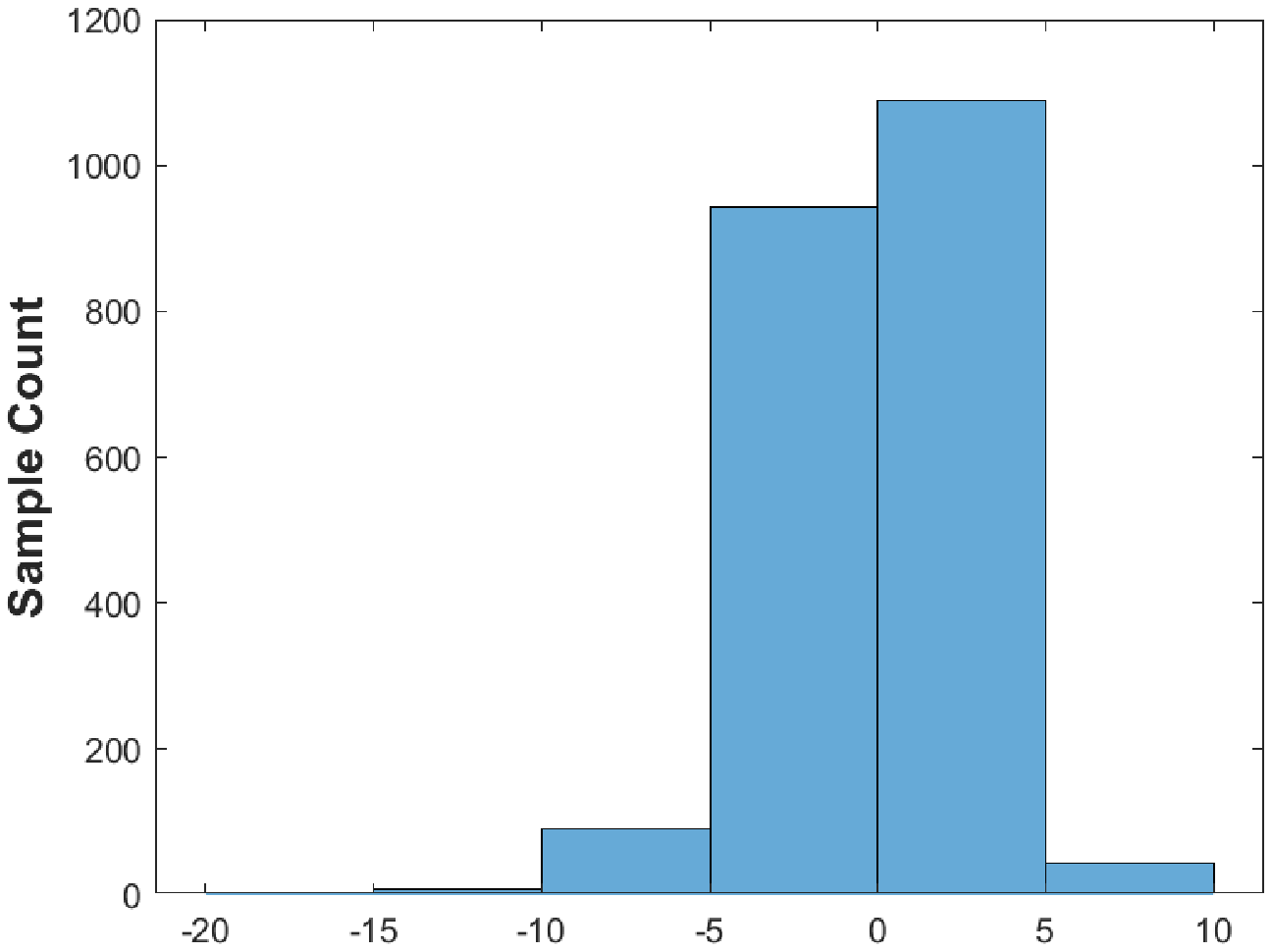} 
		\caption{\tindog{}}
		\end{subfigure}%
 \hfill
		\begin{subfigure}[t]{0.5\textwidth}
		\centering
		\includegraphics[width=6cm]{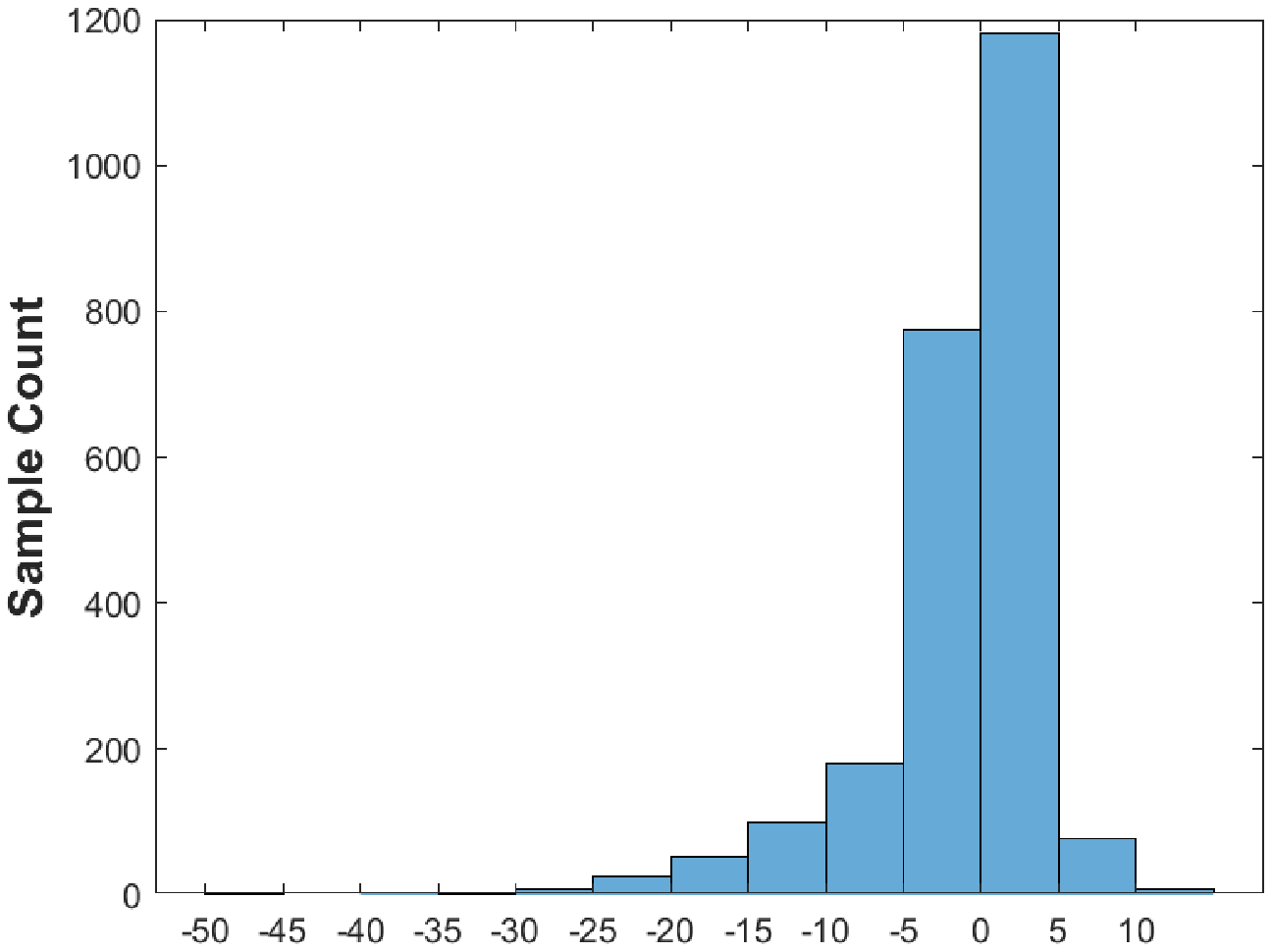} 
		\caption{\homestead{}}
		\end{subfigure}%
\caption{Histogram of the difference between laboratory Fe and SVM estimation of Fe on \tindogandhomestead{} samples}
\label{Fig007}
\end{figure*}

\begin{figure*}[t!]
		\begin{subfigure}[t]{0.5\textwidth}
		\centering
		\includegraphics[width=6cm]{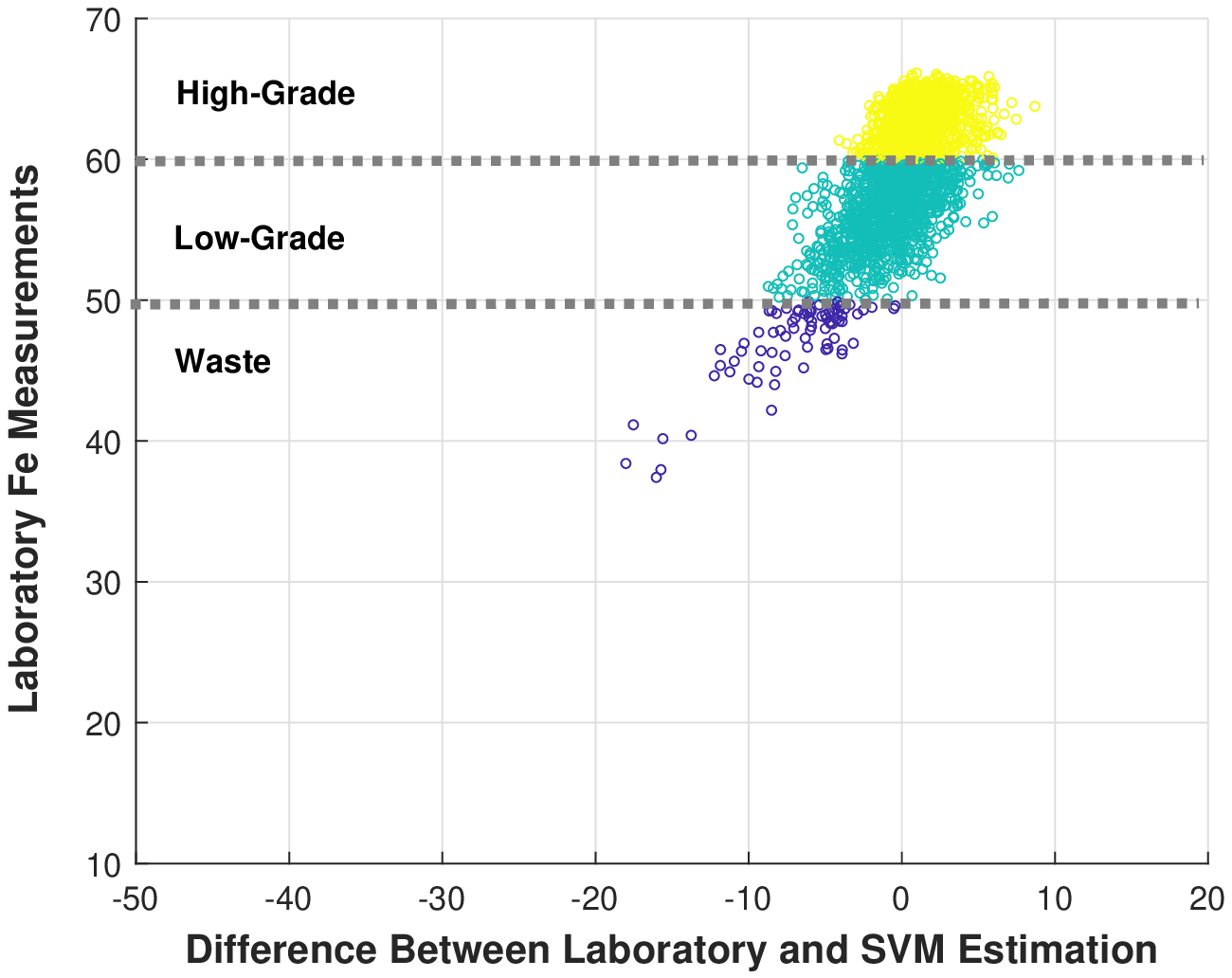} 
		\caption{\tindog{}}
		\end{subfigure}%
 \hfill
		\begin{subfigure}[t]{0.5\textwidth}
		\centering
		\includegraphics[width=6cm]{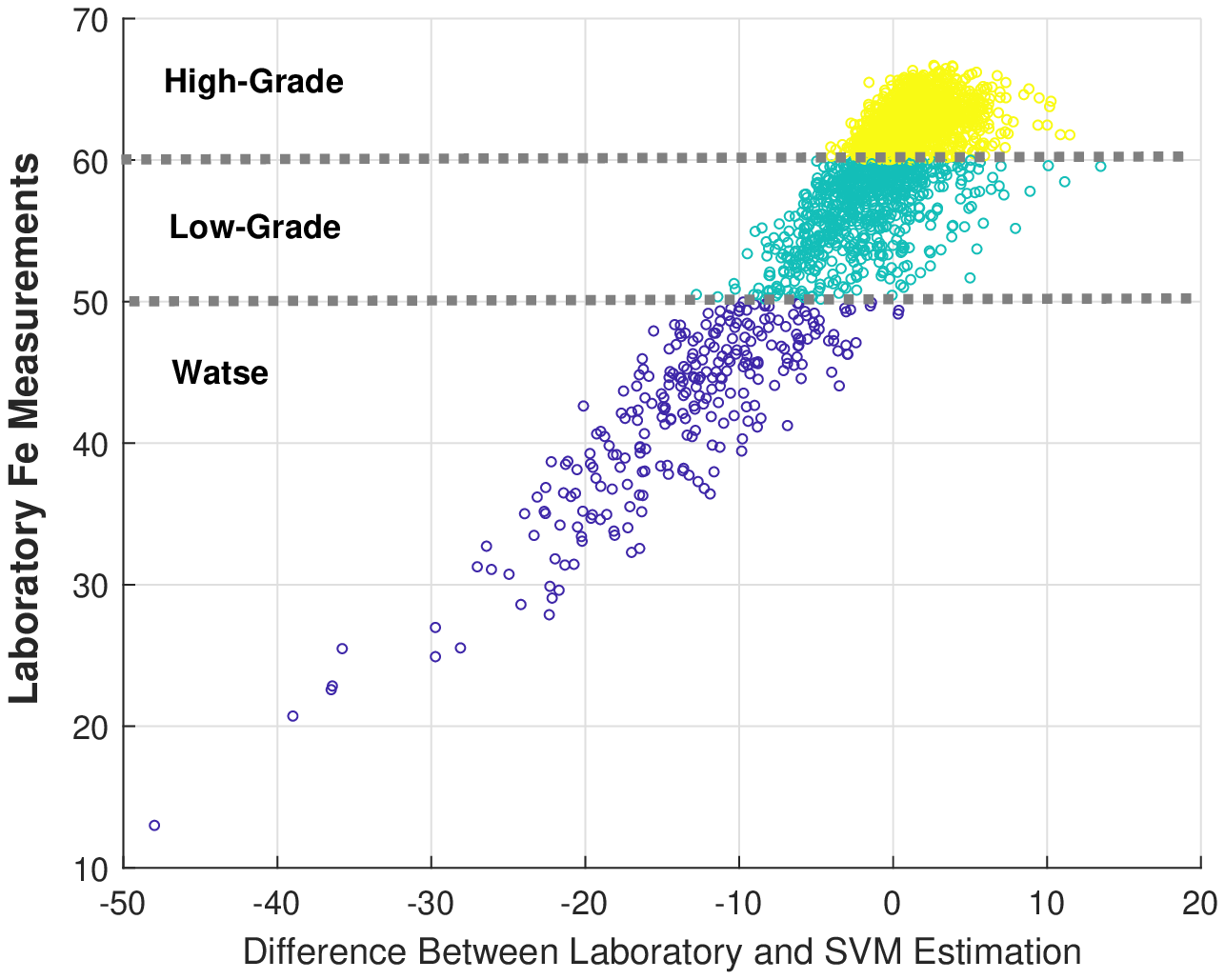} 
		\caption{\homestead{}}
		\end{subfigure}%
\caption{The difference between laboratory Fe and SVM estimation of Fe against the classifications of Fe into low, med and high-grades, on \tindogandhomestead{} samples}
\label{Fig008}
\end{figure*}

As the performance of these models on the Fe data was similar across both mining sites, the estimates generated by the SVM models were selected for further analysis. The histogram of the residuals, that is the difference between laboratory Fe measurements and SVM Fe estimations, are shown in \figurename{ \ref{Fig007}}. These histograms clearly indicate that the majority of \tindog{} samples had a difference between laboratory and estimated values of $\pm 2.5$, which is the typical tolerance in the assay error for Fe and SiO$_2$ as indicated in the literature \citep{DanielAVA}. However, the \homestead{} residuals were concentrated around $\pm 5$. It is important to note here that the tolerances are usually set independently for each element and may vary from project to project, according to requirements or the interval’s iron grade, as less accurate validation is required for low-grade (waste) intervals. In general, low Fe values of Fe $<$ 50 are classified into waste, while $50 \leq \textrm{Fe} < 60$ are classified as med-grade and Fe $\geq$ 60 are classified as high-grade. When plotting the residuals across these three classes of Fe grades, it was observed that the tail of the histograms in \figurename{ \ref{Fig006}} with large negative values were all related to low-grade iron as shown in \figurename{ \ref{Fig008}}. These larger errors occur well under the waste cut-off, so are unimportant as the exact Fe content will not be considered in further decision-making at such low values.

\subsection{Phosphorus (P) Prediction}
\begin{figure}[t!]
        \begin{subfigure}[t]{\textwidth}
        \centering
		    \includegraphics[width=10cm,trim={2cm 1cm .5cm 1cm},clip]{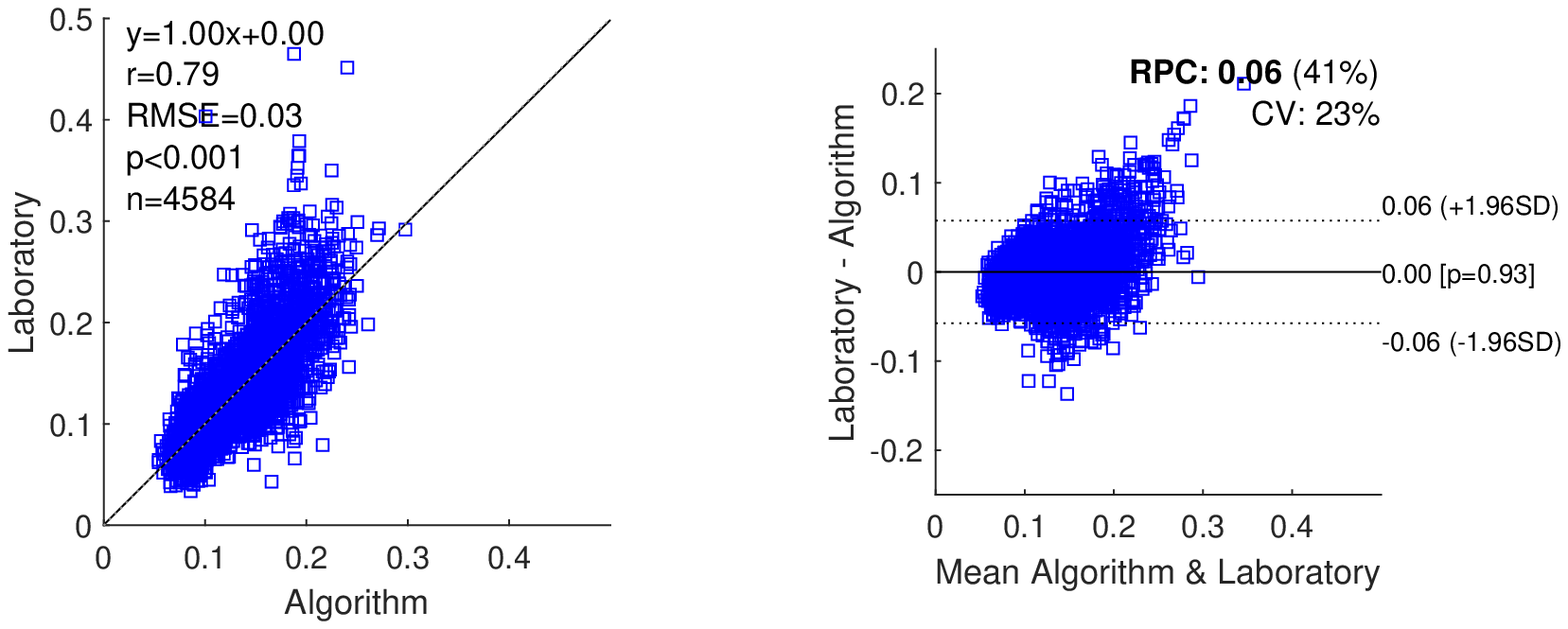} 
		    \caption{GP model results}
    \end{subfigure}%
   
		\begin{subfigure}[t]{\textwidth}
        \centering
         \includegraphics[width=10cm,trim={2cm 1cm .5cm 1cm},clip]{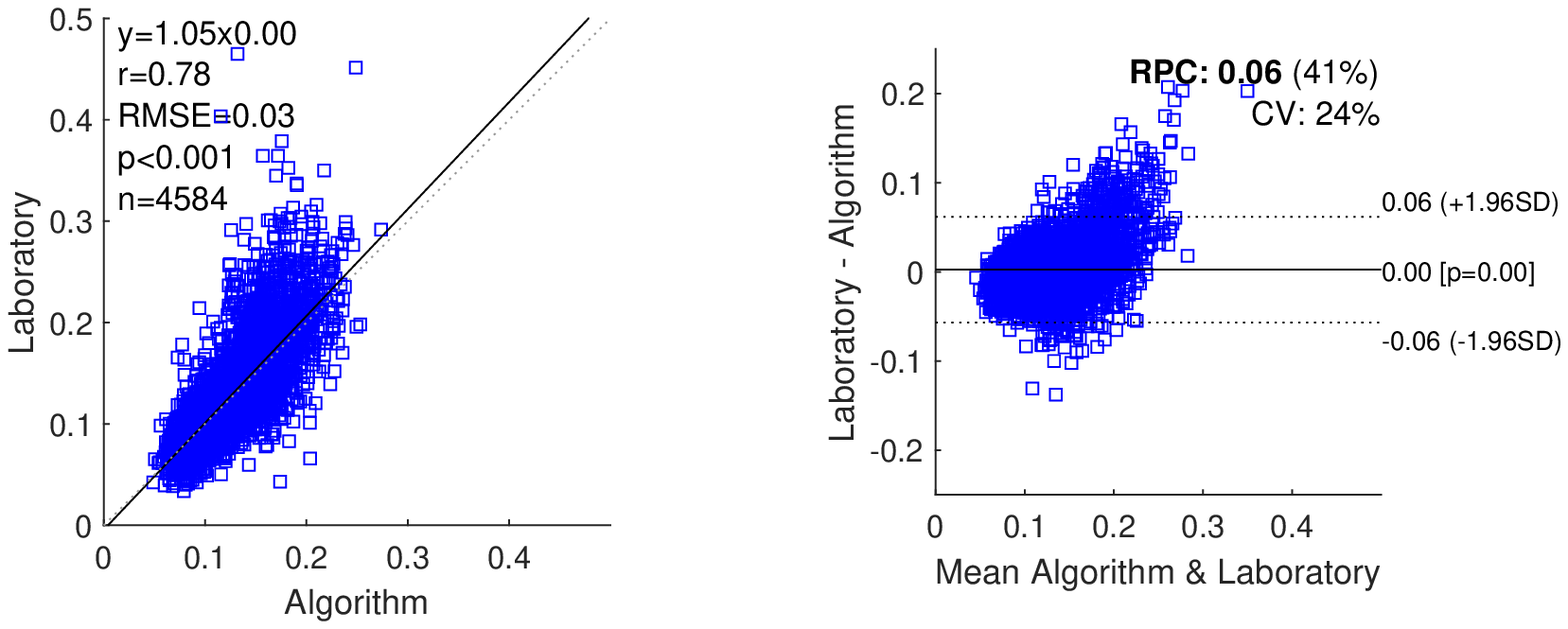} 
		     \caption{SVM model results}
    \end{subfigure}%
	
	\begin{subfigure}[t]{\textwidth}
        \centering
         \includegraphics[width=10cm,trim={2cm 1cm .5cm 1cm},clip]{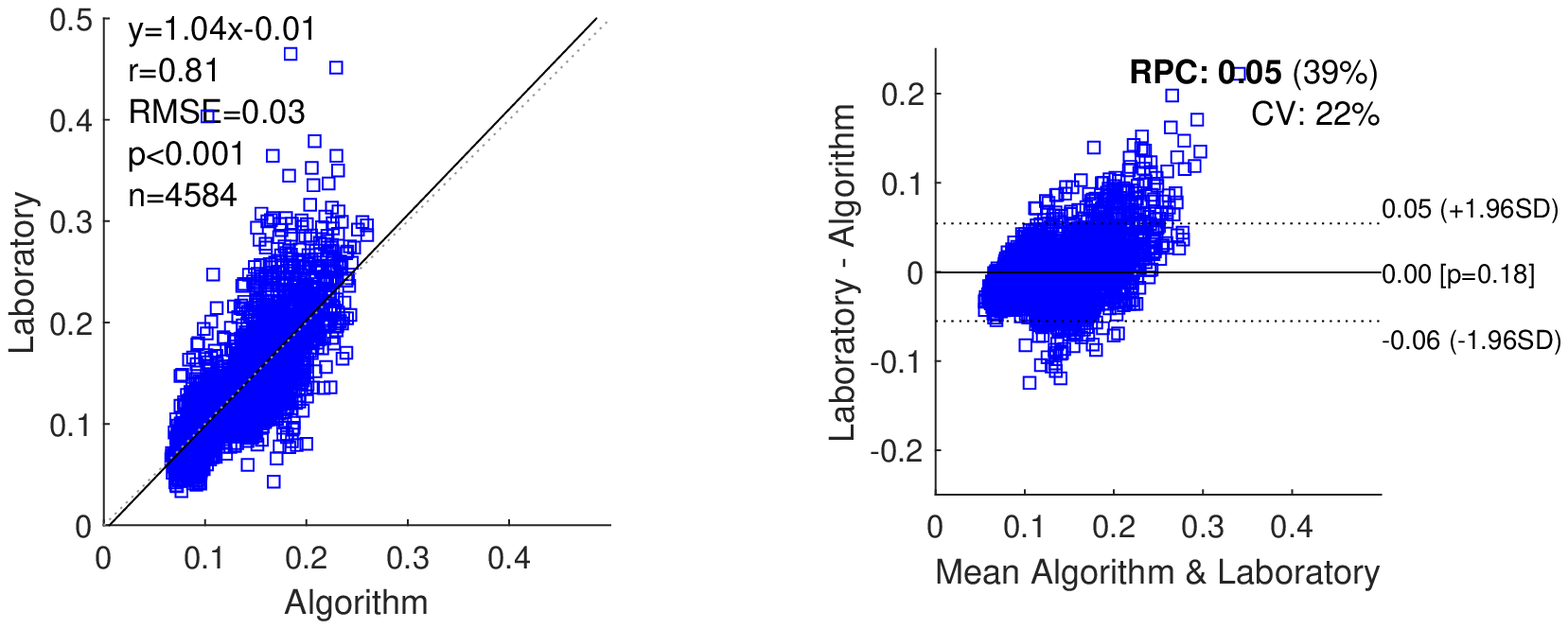} 
		     \caption{RF model results}
    \end{subfigure}%
    
    \caption{Combined Phosphorus (P) data, \tindogandhomestead{}, Bland-Altman regression plots with several machine learning models}
		\label{Fig009}
\end{figure}

In these experiments, the two datasets from \tindog{} and \homestead{} were combined to generate one large set of samples, as each site contains a relatively small number of samples. The results in \figurename{ \ref{Fig009}} demonstrate the performance of the machine learning models in predicting P values from the MWD data. In this case, the RF model demonstrated the highest correlation with the laboratory measurements of 0.81, with an RMSE of 0.03. The GP model had a correlation coefficient of 0.79, and SVM of 0.78, both with an RMSE of 0.03.

\begin{figure}[h]
	\centering
	\includegraphics[width=0.5\textwidth]{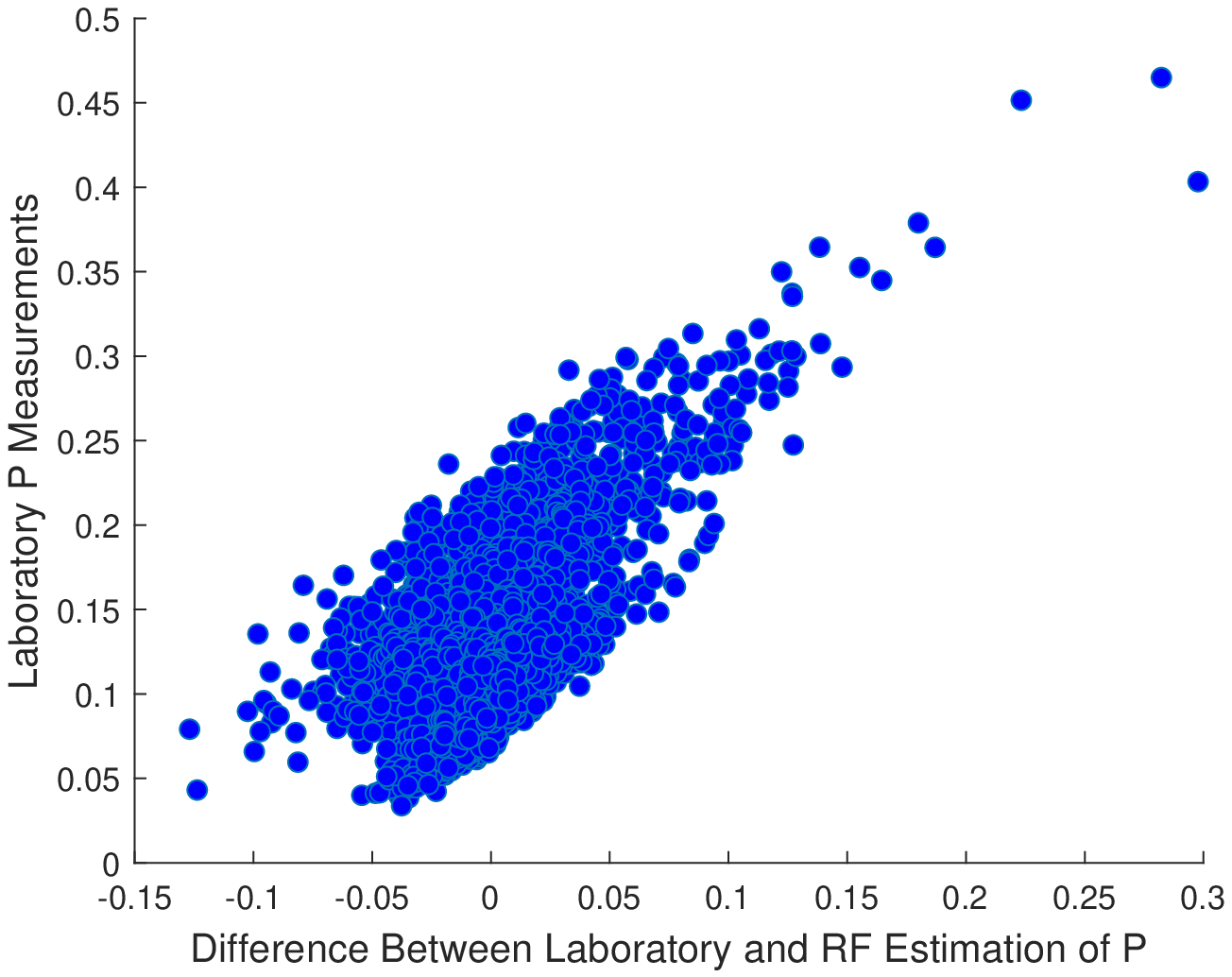}
	\caption{Difference between laboratory measurements and RF estimation of P vs. laboratory P measurements}
	\label{Fig010}
\end{figure}

In order to understand more about what impacted these correlation values, we plotted the difference between the laboratory P measurements and RF predictions of P versus the laboratory P measurements. In this way, one can clearly see where the laboratory and model disagree, and observe the actual range of P across the corresponding samples. The results in \figurename{ \ref{Fig010}} clearly show that very few of the approximately 4000 samples had a measured P value $> 0.35$, and these were essentially `missed' by the model, as the model did not observe sufficient samples of this magnitude to learn the relationship between MWD and P across that range. Hence, the sample size available for this experiment did impact the accuracy of estimation. It is believed that having much larger datasets with a significantly larger number of samples would enhance the models' performance. Considering the available samples, the performance of the models with an RMSE of 0.03 is quite promising, given that the literature has no previously reported figures in this regard, to the best of the authors' knowledge.

\subsection{Sulfur (S) Prediction}
\begin{figure}[t!]
        \begin{subfigure}[t]{\textwidth}
        \centering
		    \includegraphics[width=10cm,trim={2cm 1cm .5cm 1cm},clip]{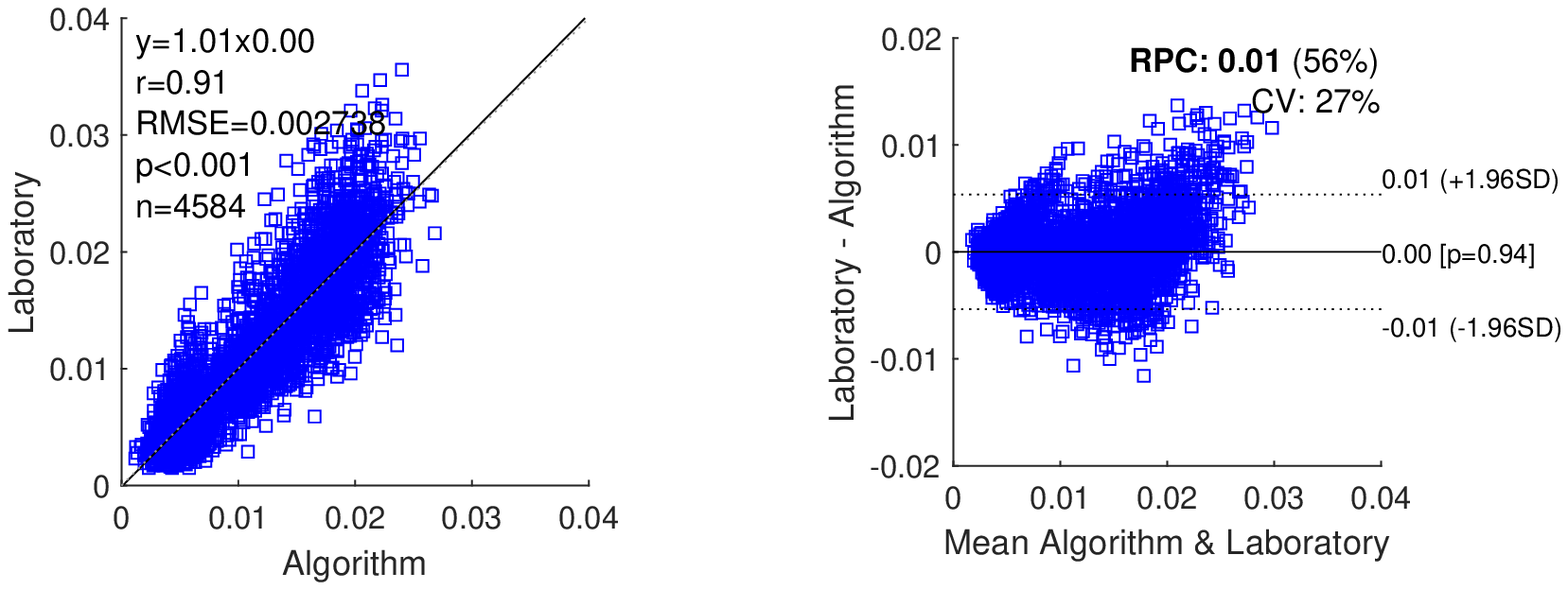} 
		    \caption{GP model results}
    \end{subfigure}%
   
		\begin{subfigure}[t]{\textwidth}
        \centering
         \includegraphics[width=10cm,trim={2cm 1cm .5cm 1cm},clip]{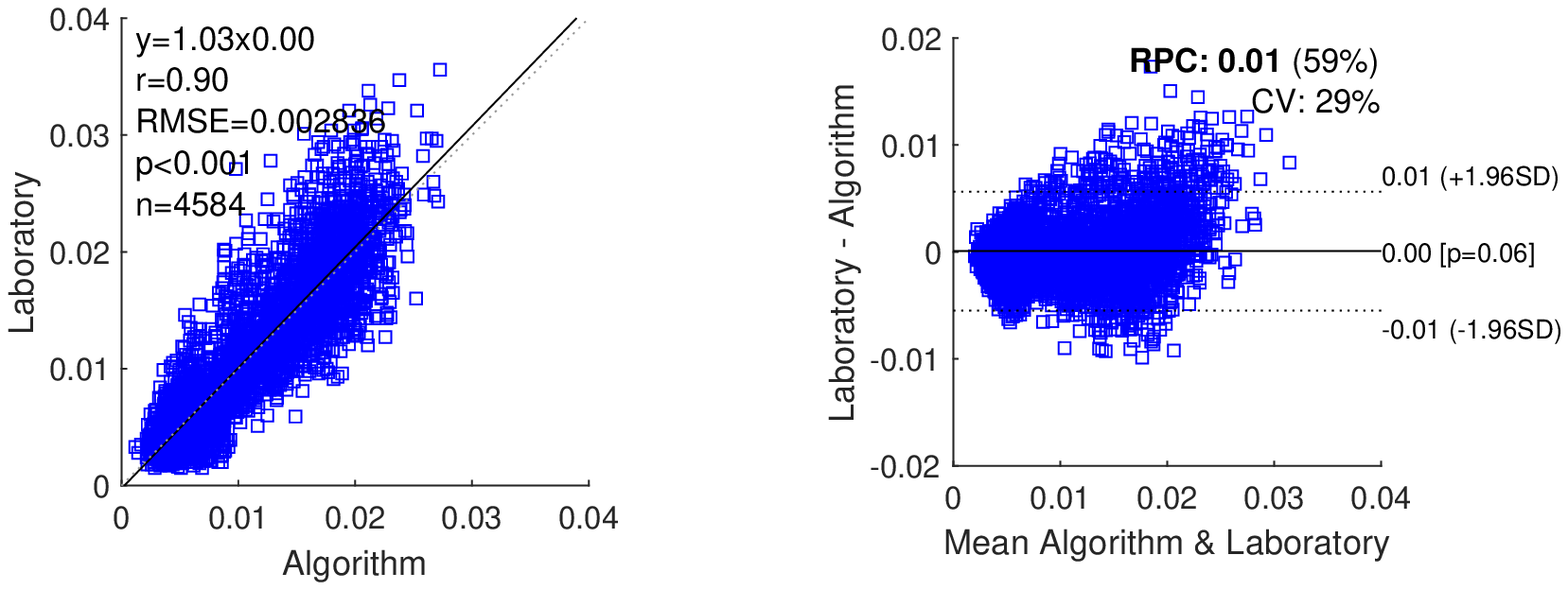} 
		     \caption{SVM model results}
    \end{subfigure}%
	
	\begin{subfigure}[t]{\textwidth}
        \centering
         \includegraphics[width=10cm,trim={2cm 1cm .5cm 1cm},clip]{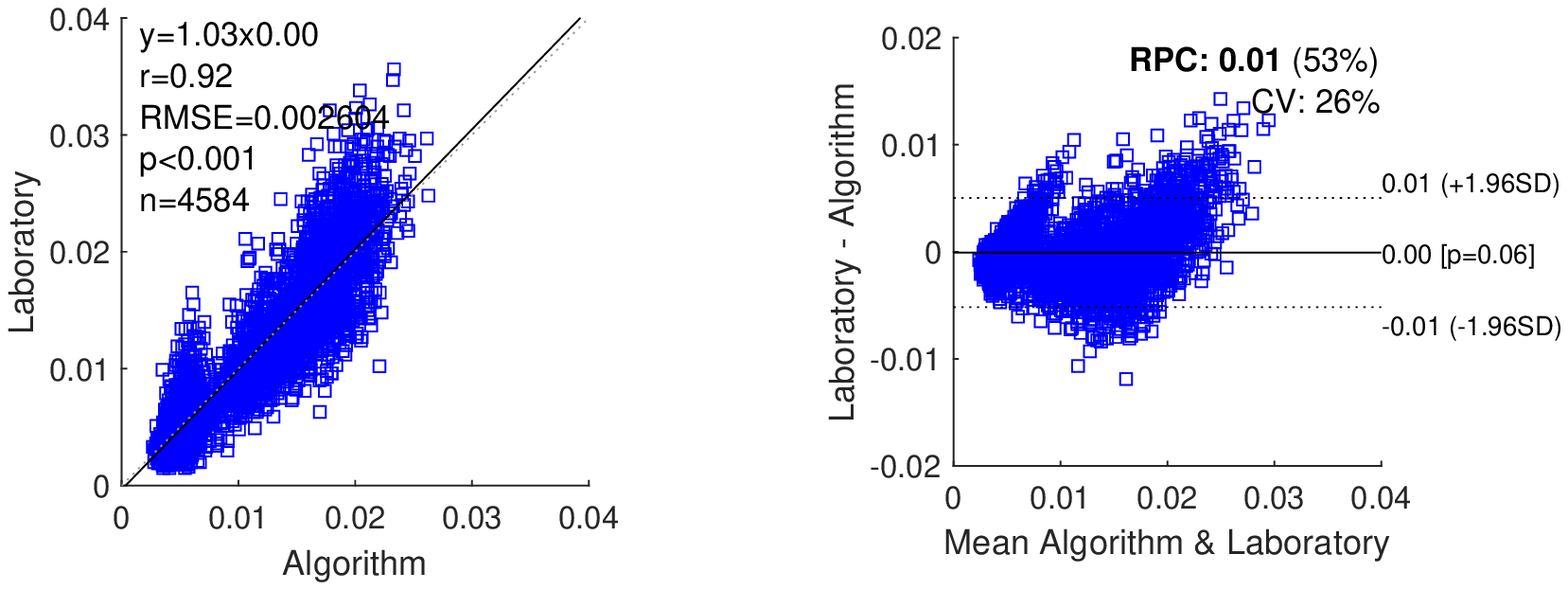} 
		     \caption{RF model results}
    \end{subfigure}%
    
    \caption{Combined Sulfur (S) data, \tindogandhomestead{}, Bland-Altman regression plots with several machine learning models}
		\label{Fig011}
\end{figure}

\begin{figure}[h]
	\centering
	\includegraphics[width=.75\textwidth]{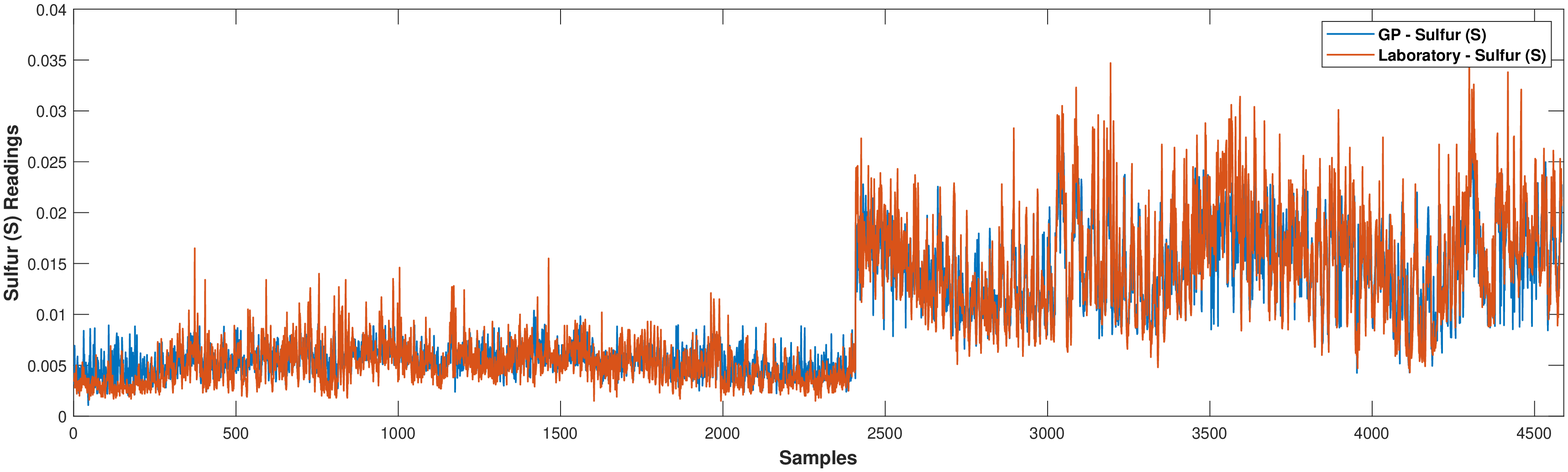}
	\caption{Raw Sulfur (S) laboratory and GP model predictions showing high agreement between measurements and predictions}
	\label{Fig012}
\end{figure}

The results for predicting S based on the combined MWD data from \tindogandhomestead{} are shown in \figurename{ \ref{Fig011}} using the GP, SVM, and RF models. These results clearly demonstrate that the accuracy of the proposed approach generally depends on the assay type (for example, Fe, P, S, etc.) and distribution within the two mining sites, with Sulfur having correlation coefficients as high as 0.91 using GP model with an RMSE of 0.0027. On the other hand, the raw measurements of S and their GP predictions based on MWD data are also shown in \figurename{ \ref{Fig012}}, which further visually demonstrates the effectiveness of the method proposed in this paper.

\subsection{Spatially-Aware Cross-Validation}

\begin{figure*}[t!]
\centering
    \begin{subfigure}[t]{0.48\columnwidth}
				\centering
        \includegraphics[width=7cm]{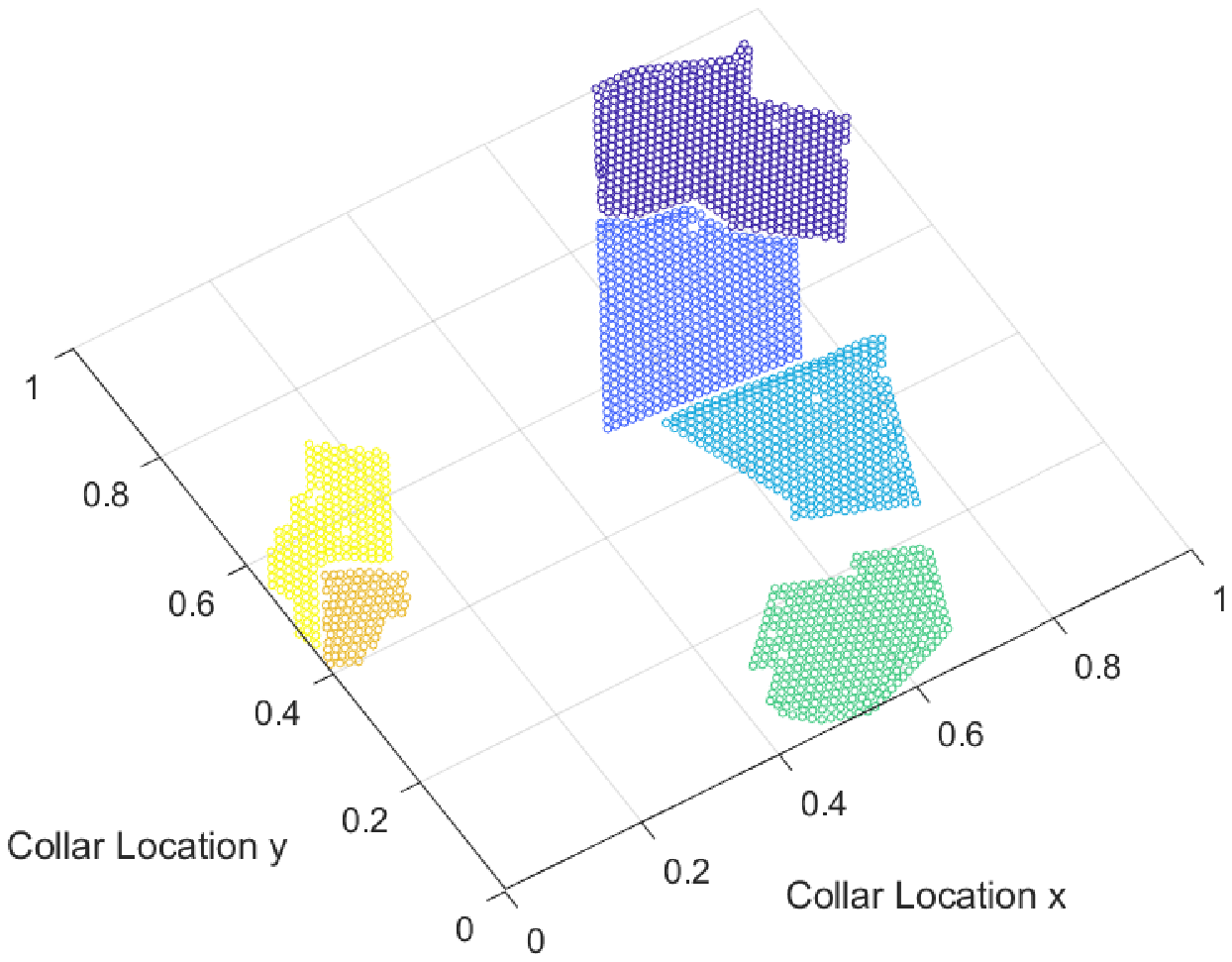} 
		    \caption{\tindog{} blast-hole distribution}
    \end{subfigure}%
   \hfill
		\begin{subfigure}[t]{0.48\columnwidth}
				 \centering
         \includegraphics[width=7cm]{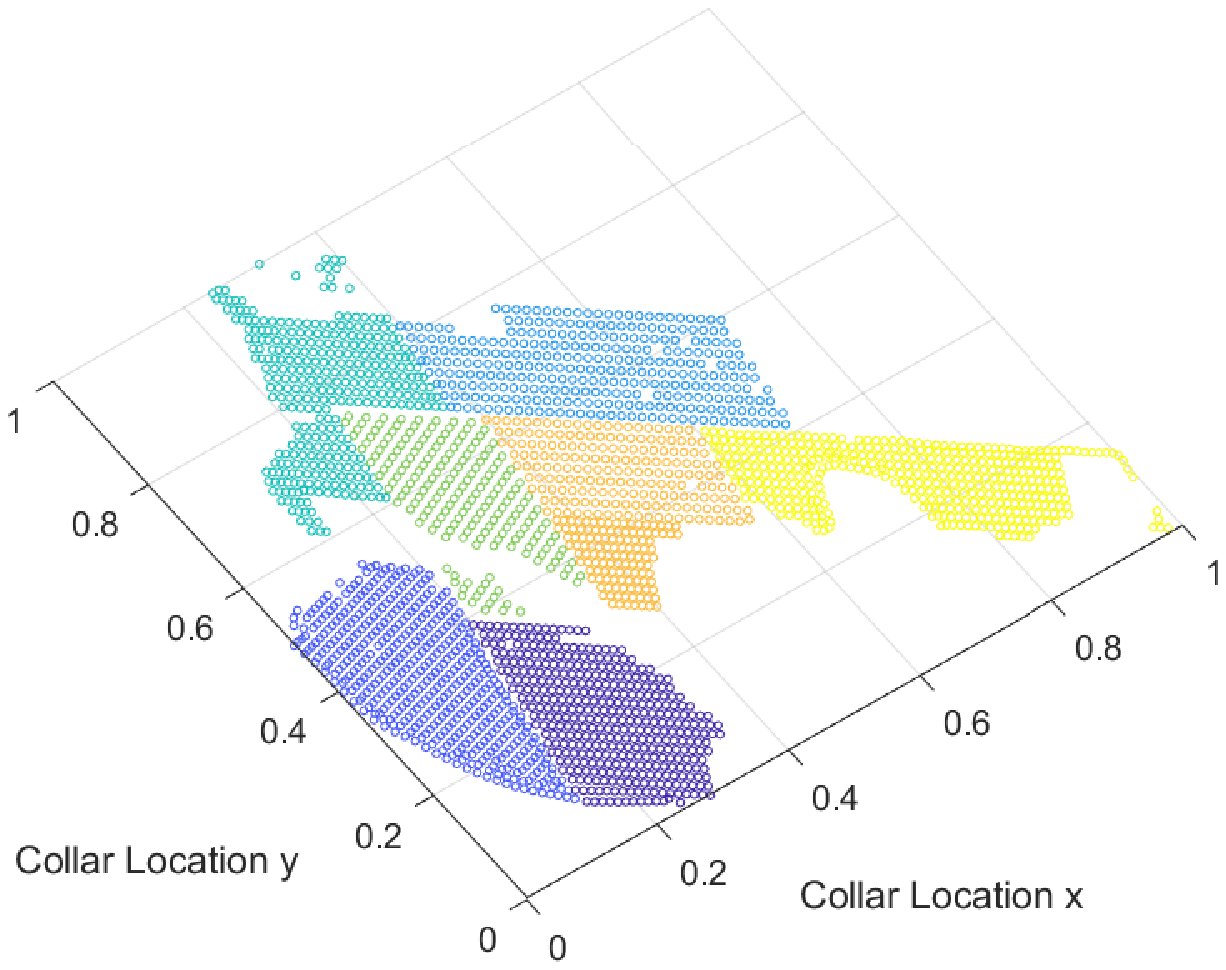} 
		     \caption{\homestead{} blast-hole distribution}
    \end{subfigure}%
	
	  \caption{Spatial distribution of the blast-holes across the two sites, with colors representing the different blasts across the two sites. The x-y coordinates have been re-scaled due to commercial sensitivity}
		\label{FigSpatial}
\end{figure*}

In order to develop a spatially-aware cross-validation scheme, we first plot the spatial distribution of the available blast-holes across the patterns from the two mining sites of \tindog{} and \homestead{} as shown in \figurename{ \ref{FigSpatial}}. Our spatial cross-validation process involved a leave-one-blast-out process, in which all of the holes related to an individual blast are kept for testing while training is based on the data from all other holes from all remaining blasts. The process is repeated across each of the blasts to ensure that the testing is performed across all of the blasts in the data set. Performance estimates are computed based on the final results to validate the overall assay estimation performance across spatially different blasts.     

\begin{figure}[t!]

    \begin{subfigure}[t]{\linewidth}
        \centering
		    \includegraphics[width=10cm,trim={2cm 1cm .5cm 1cm},clip]{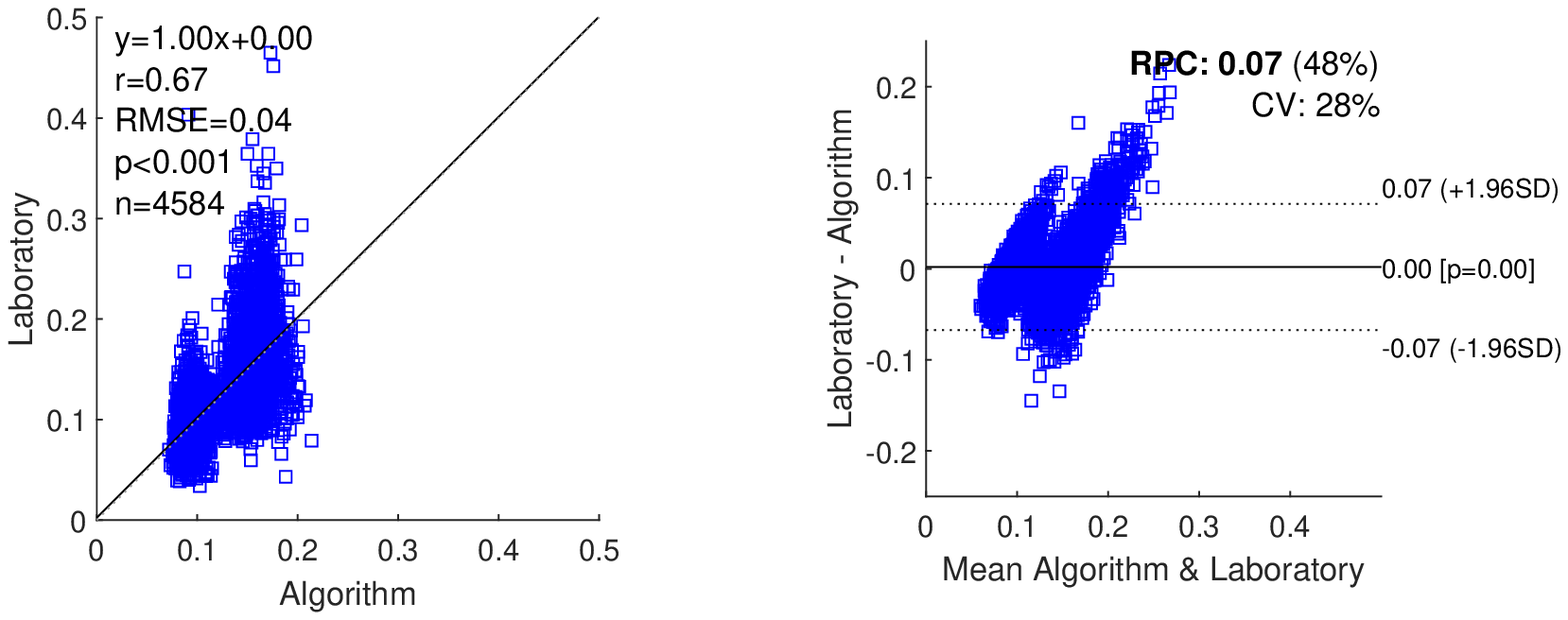} 
		    \caption{Phosphorus (P) regression results}
    \end{subfigure}%
   
		\begin{subfigure}[t]{\linewidth}
        \centering
         \includegraphics[width=10cm,trim={2cm 1cm .5cm 1cm},clip]{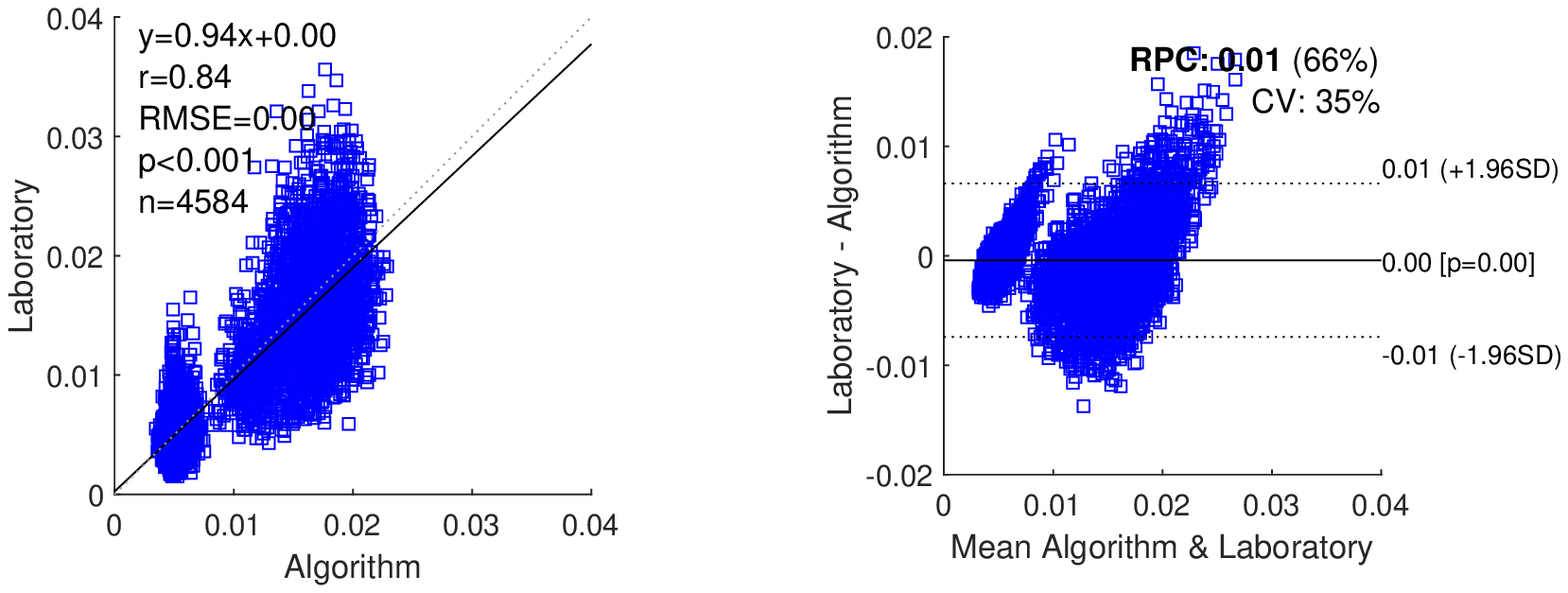} 
		     \caption{Sulfur (S) regression results}
    \end{subfigure}%
	
	  \caption{Spatially-aware regression results for Phosphorus (P) and Sulfur (S) using individual RF models}
		\label{FigSpatial_P_S}
\end{figure}

The spatially-aware regression results for both P and S are shown in \figurename{\ref{FigSpatial_P_S}} using an RF model. These results demonstrate that the model's performance when estimating these two chemical assays using a spatially-aware mode significantly dropped in comparison to the previously demonstrated P and S results when using random k-fold cross-validation. This is in line with the literature findings in that, in the random k-fold cross-validation mode, autocorrelation could have impacted the results. Given the purpose of our first experiment was to fill in the missing gaps of chemical assays for blast-holes, the previous performance measures are advantageous within the context and objective of this first testing scheme. In comparison, the results of the spatially-aware cross-validation scheme could have been impacted by several factors including first, the limited size of the available data, with only 4584 samples across both sites, and second, the different distributions of the assays across the two sites. Hence, the authors conclude that further experimentation with a significantly larger dataset would be required to properly judge the modelling accuracy of assays from MWD across spatially distance locations.

\subsection{Cross-Assay Predictions}
\begin{figure}[h]
	\centering
	\includegraphics[width=.5\textwidth]{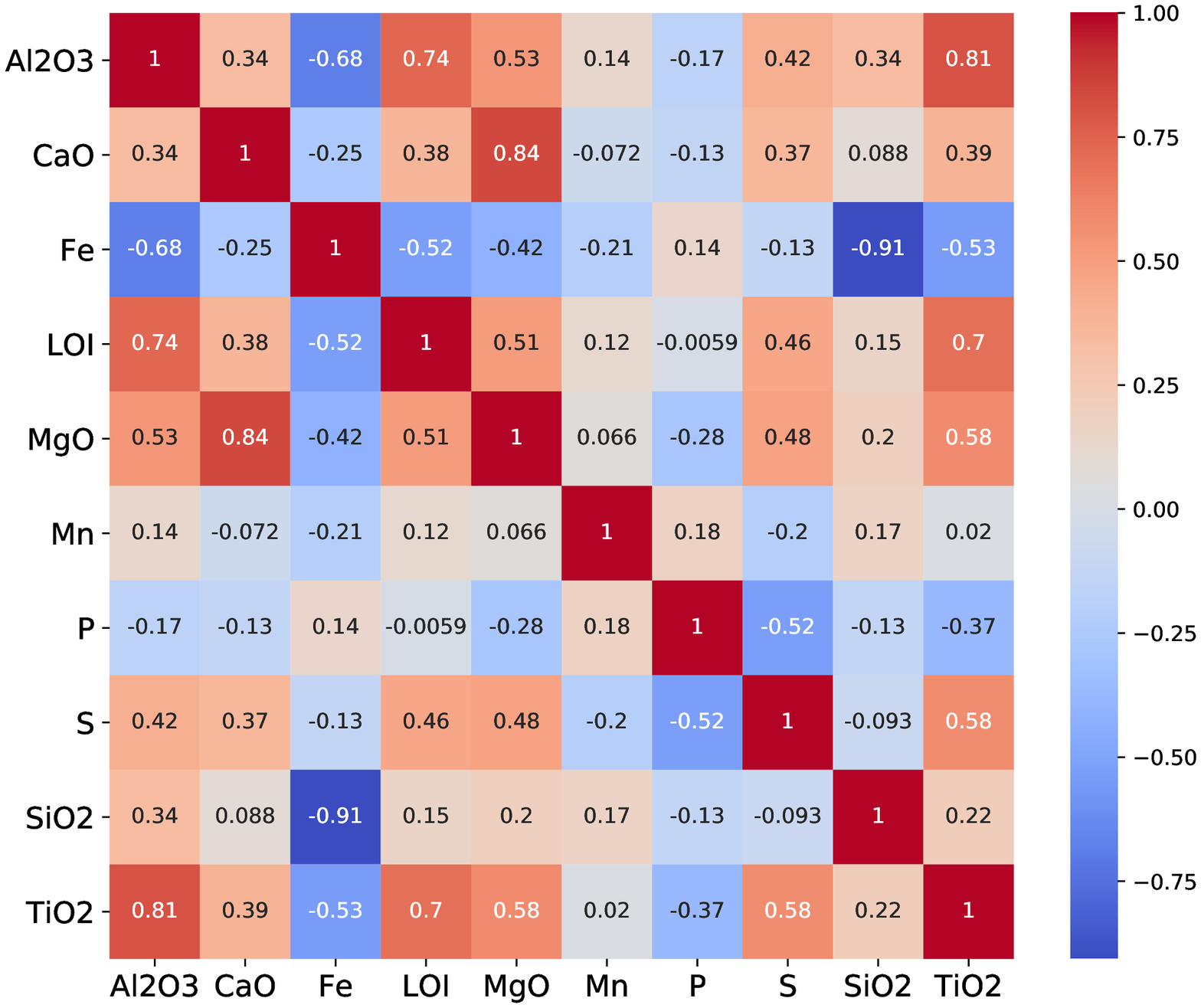}
	\caption{Pearson correlation coefficients among all assays from both mining sites}
	\label{Fig013}
\end{figure}

\begin{figure}[t!]
        \begin{subfigure}[t]{\textwidth}
        \centering
		    \includegraphics[width=10cm,trim={2cm 1cm .5cm 1cm},clip]{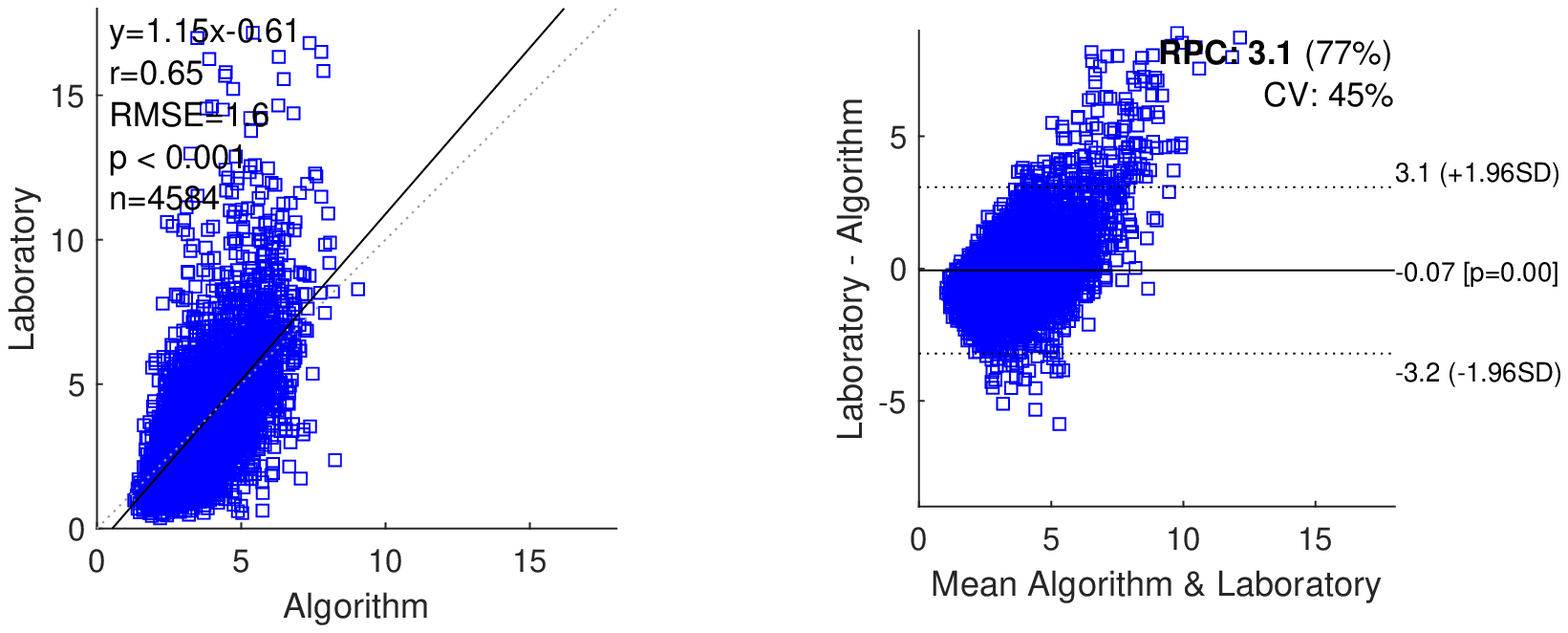} 
		    \caption{RF model results predicting Al\textsubscript{2}O\textsubscript{3} using MWD only}
    \end{subfigure}%
   
		\begin{subfigure}[t]{\textwidth}
        \centering
         \includegraphics[width=10cm,trim={2cm 1cm .5cm 1cm},clip]{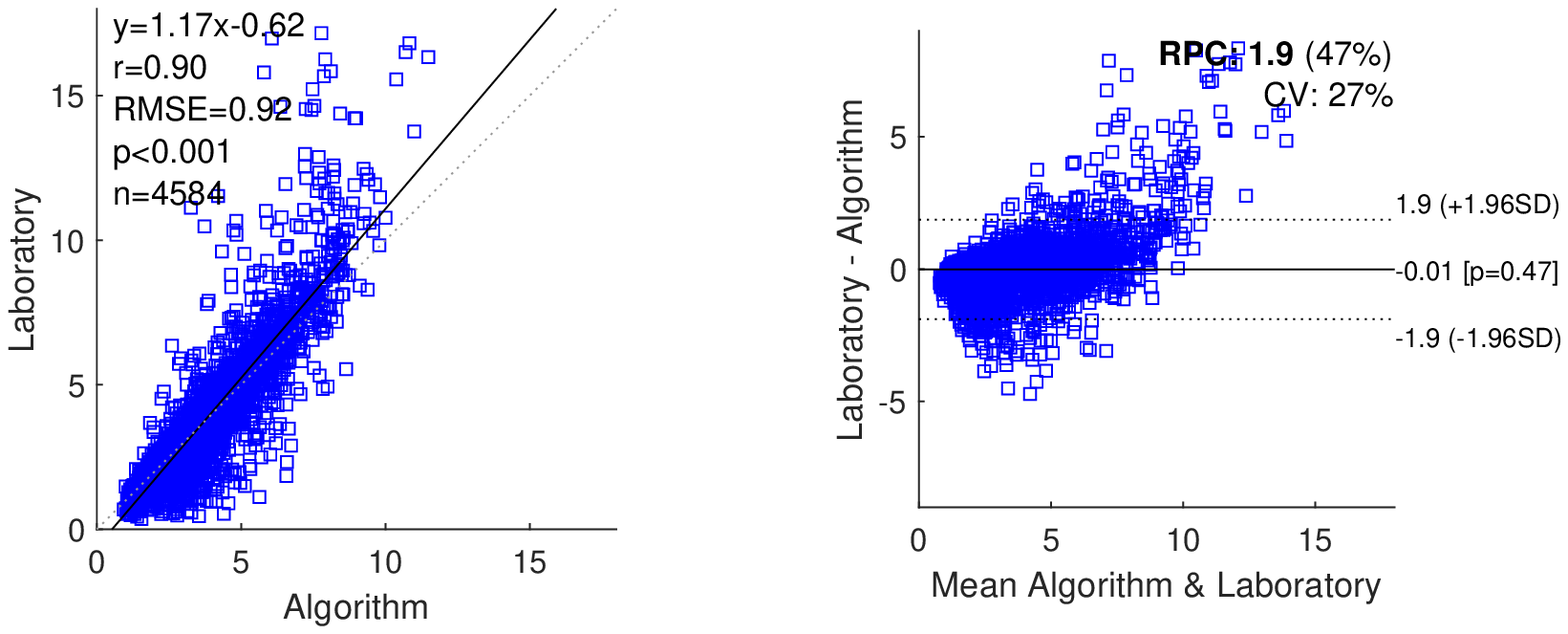} 
		     \caption{RF model results predicting Al\textsubscript{2}O\textsubscript{3} using MWD with Fe}
    \end{subfigure}%
	
	\begin{subfigure}[t]{\textwidth}
        \centering
         \includegraphics[width=10cm,trim={2cm 1cm .5cm 1cm},clip]{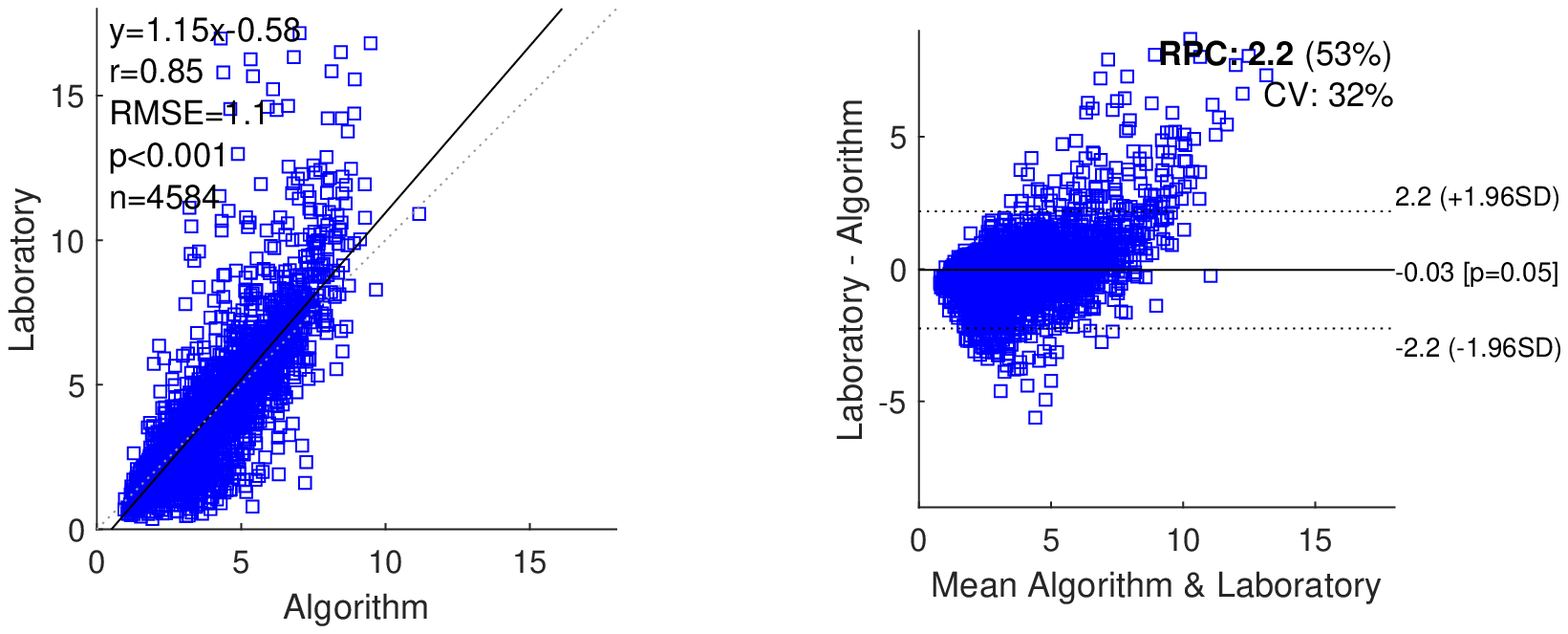} 
		     \caption{RF model results predicting Al\textsubscript{2}O\textsubscript{3} using MWD with SiO\textsubscript{2}.}
    \end{subfigure}%
    
    \caption{The impact of combining MWD with other individual chemistry assays to predict Al\textsubscript{2}O\textsubscript{3}}
		\label{Fig014}
\end{figure}

The goal of this experiment is to verify whether the knowledge about one assay type can better inform the models about the distributions of other assays, that is to indicate if, for example, the knowledge of Fe or SiO\textsubscript{2} can inform Al\textsubscript{2}O\textsubscript{3}, or other such assay combinations. For this specific test, the RF model was selected as an example of the available regressions models, as the performance is similar across GP, SVM and RF, and the purpose is simply to demonstrate the concept. The experiment proceeded by first analyzing whether there is any correlation between the different assays, which is shown to be the case in \figurename{ \ref{Fig013}}. The results for predicting Al\textsubscript{2}O\textsubscript{3} using MWD alone, MWD together with Fe, and MWD together with SiO\textsubscript{2} are each shown in \figurename{ \ref{Fig014}}. These results show significant enhancements to Al\textsubscript{2}O\textsubscript{3} accuracy (correlation and RMSE) when using the combination of MWD with one additional chemistry assay. This supports the hypothesis that knowing one assay type can significantly enhance the prediction performance for other assay types. The benefit of this finding is that if one assays type can be measured or predicted with some sensor/model, then these measurements can be used to improve estimates for other assays, a subject of future investigation. 

\begin{figure}[t!]

    \begin{subfigure}[t]{\textwidth}
        \centering
		    \includegraphics[width=10cm,trim={2cm 1cm .5cm 1cm},clip]{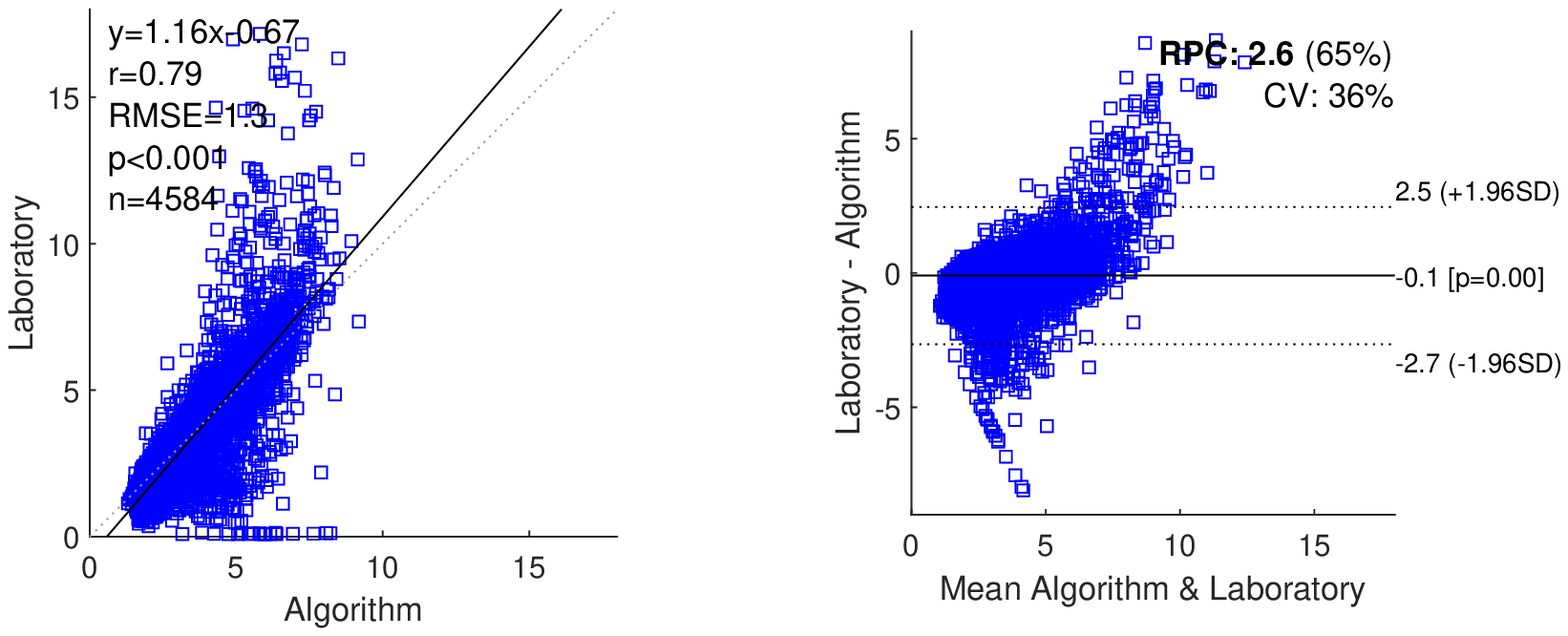} 
		    \caption{RF model results predicting Al\textsubscript{2}O\textsubscript{3} using MWD with Fe from a spatially-aware cross-validation scheme}
    \end{subfigure}%
   
		\begin{subfigure}[t]{\textwidth}
        \centering
         \includegraphics[width=10cm,trim={2cm 1cm .5cm 1cm},clip]{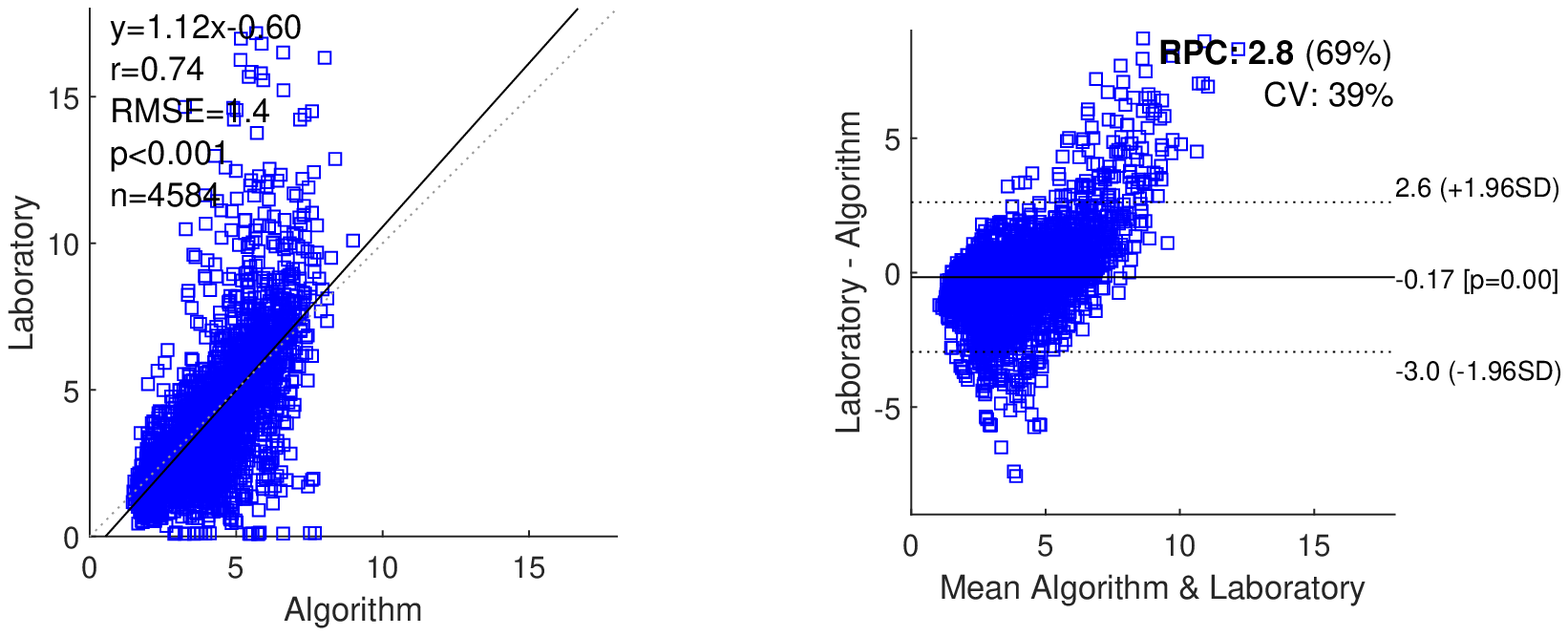} 
		     \caption{RF model results predicting Al\textsubscript{2}O\textsubscript{3} using MWD with SiO\textsubscript{2} from a spatially-aware cross-validation scheme}
    \end{subfigure}%
	
	  \caption{Spatially-aware cross-validation test on cross-assay prediction}
		\label{FigSpatialBetweenAssays}
\end{figure}

The analysis also included a spatially-aware cross-validation test on cross-assay prediction. In this regard, the results for predicting Al\textsubscript{2}O\textsubscript{3} using MWD together with Fe, and MWD together with SiO\textsubscript{2} are shown in \figurename{ \ref{FigSpatialBetweenAssays}}. It is evident from these results that the correlation coefficients dropped in comparison to those obtained by the same models with a random cross-validation scheme, which indicates the impact of autocorrelation in this case. However, these results are also encouraging as the spatially-aware models' performance can be largely enhanced by knowing the distribution of other assays estimates. For example, the same approach can be used with a technology relying on PFTNA that estimates most assays apart from P, in order to estimate these missing P values. Combining MWD with a selection of PFTNA-estimated assays to fill in the missing chemistry values warrants further investigation.

\subsection{Multivariate vs. Univariate Response Predictors} 
\begin{figure}[ht]
\centering
	\includegraphics[width=12cm]{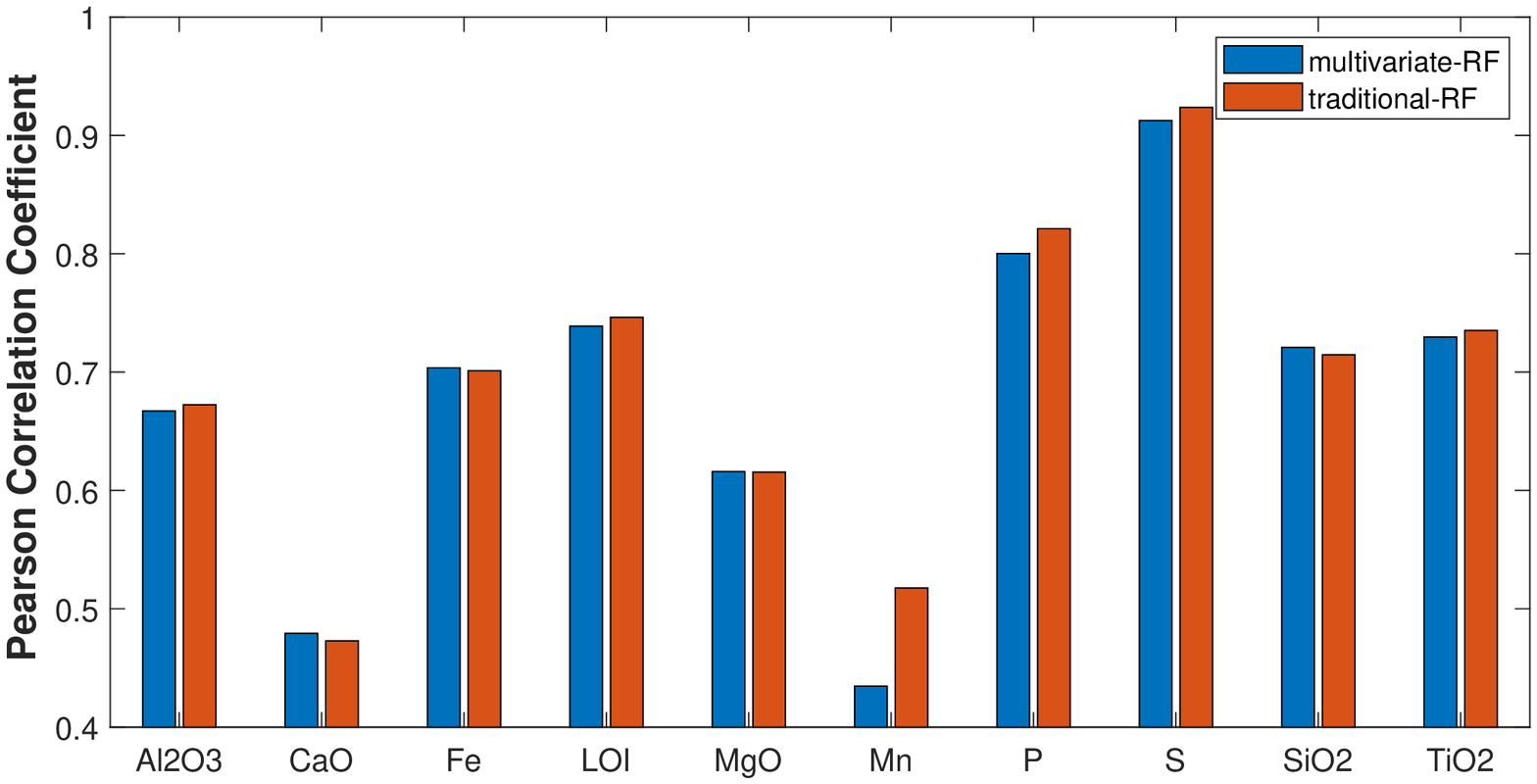}
	\caption{Multivariate vs. univariate RF response predictors}
	\label{RFverRF}
\end{figure}

While the main aim of this paper was to prove the concept of assay and material prediction via machine learning models, it could be argued that using machine learning approaches to predict multivariate response values are now available (for example, multivariate random forests) \citep{MVRF}, and thus the univariate prediction of assay grades is a limited step forward. To validate the effectiveness of the multivariate RF response predictor against the traditional univariate version, a comparison between both models was performed to predict all assays types. The results in \figurename{ \ref{RFverRF}} show no significant differences in terms of the correlation coefficients between the two multivariate and univariate RF models, except for Mn, which could be due to the limited data size and the distribution of the data from the two sites. However, it is important to note that the multivariate version predicted all responses in one step. In a real-time system deployment, this may translate to a significant reduction in the computational requirements associated with building several individual chemistry assay predictors. For offline predictions, it appears both multivariate and univariate approaches are suitable.

\subsection{Material Type Classification}
In this section, an experiment employing SVM as a classifier to predict the presence or absence of different material-types was conducted. The purpose is to estimate whether or not a given percentage of the material under consideration is present at the specified location. This is a process for which the threshold utilized to define the existence/absence of a certain material can vary from one material type to another, as different materials may become operationally relevant at different proportions. For the purpose of this paper, as we are primarily interested in a proof of concept, a threshold of 0\% was selected; any value greater than zero is taken to indicate the presence of the material under investigation. The SVM model is well suited to this analysis as SVM models were originally developed for binary classification problems such as this one.

\begin{figure*}[t!]
\centering
        \begin{subfigure}[t]{0.45\textwidth}
        \centering
		    \includegraphics[width=4cm]{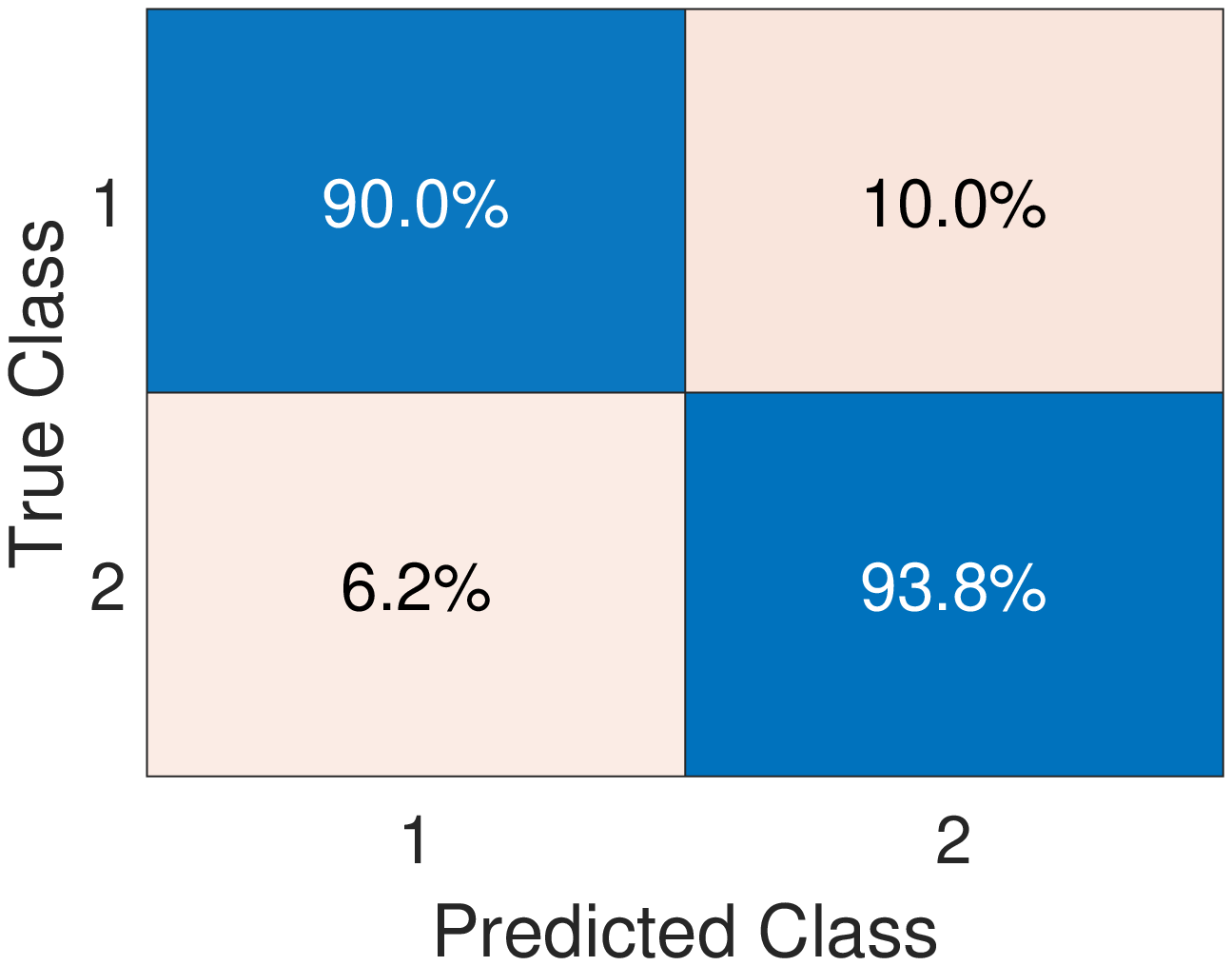} 
		    \caption{SHL confusion matrix}
    \end{subfigure}%
   ~
		\begin{subfigure}[t]{0.45\textwidth}
        \centering
         \includegraphics[width=4cm]{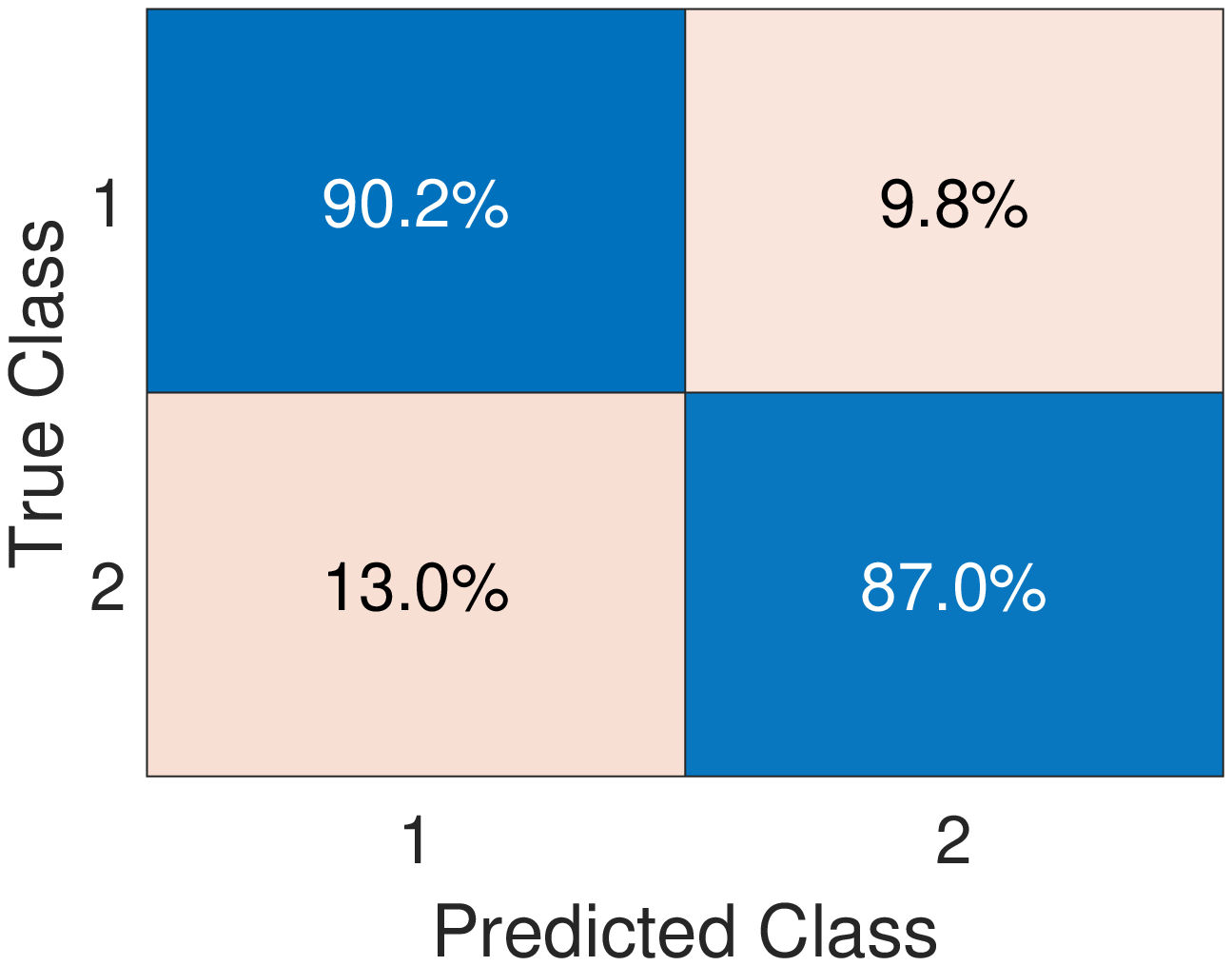} 
		     \caption{BIF confusion matrix}
    \end{subfigure}
		
		\begin{subfigure}[t]{0.45\textwidth}
        \centering
        \includegraphics[width=4cm]{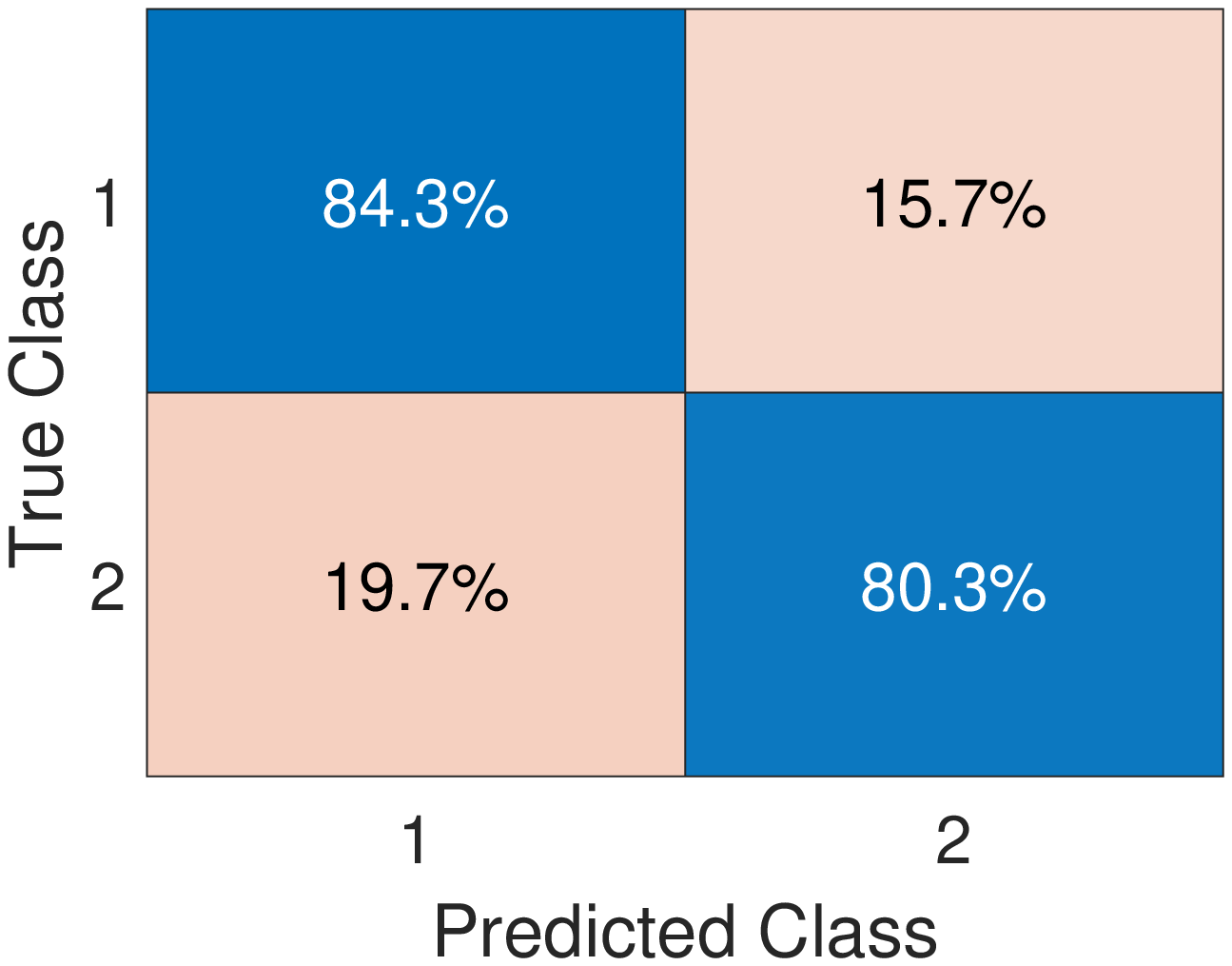} 
		    \caption{BPO confusion matrix}
    \end{subfigure}%
    ~
		\begin{subfigure}[t]{0.45\textwidth}
        \centering
         \includegraphics[width=4cm]{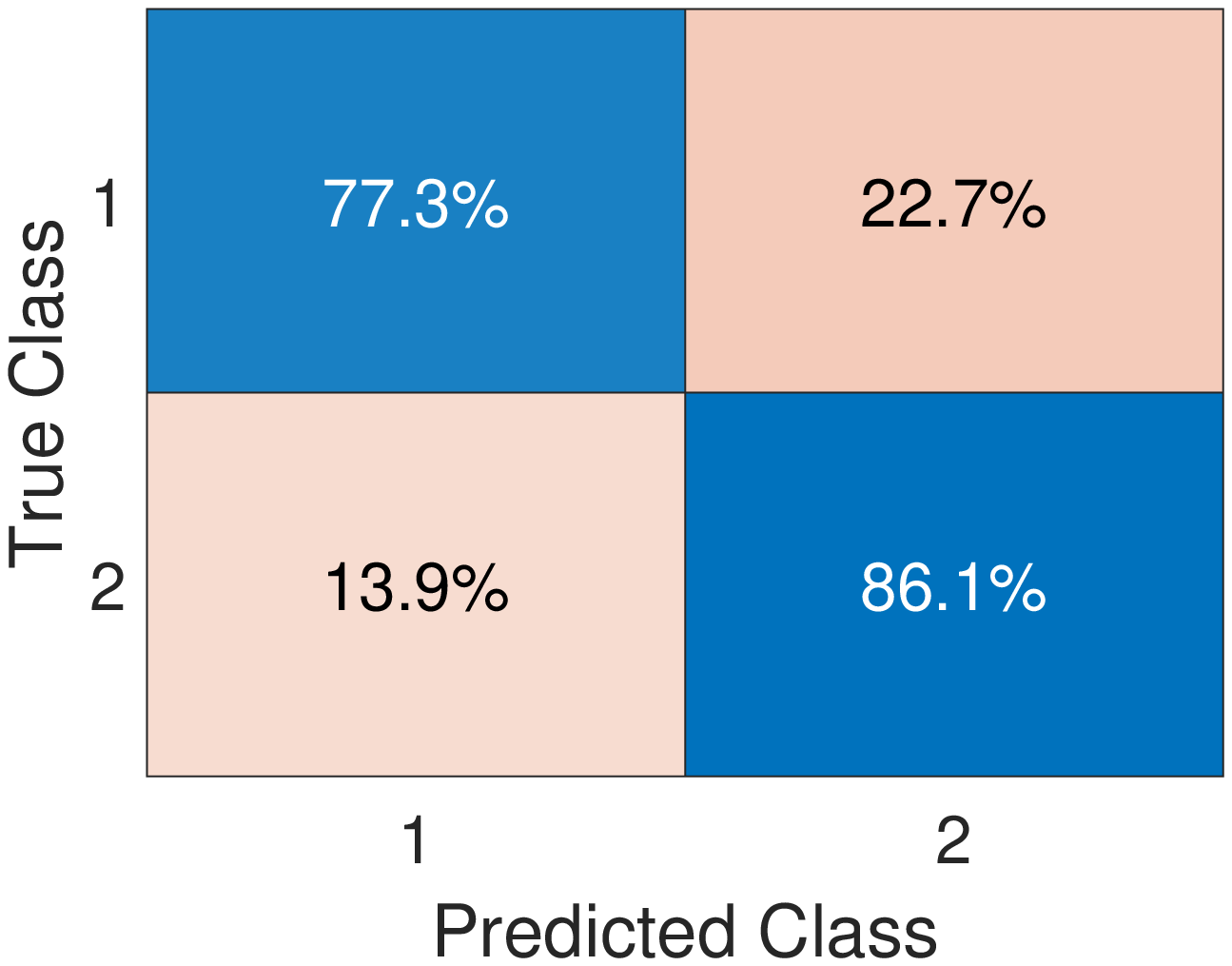} 
		    \caption{GOL confusion matrix}
    \end{subfigure}
		
		\begin{subfigure}[t]{0.45\textwidth}
        \centering
         \includegraphics[width=4cm]{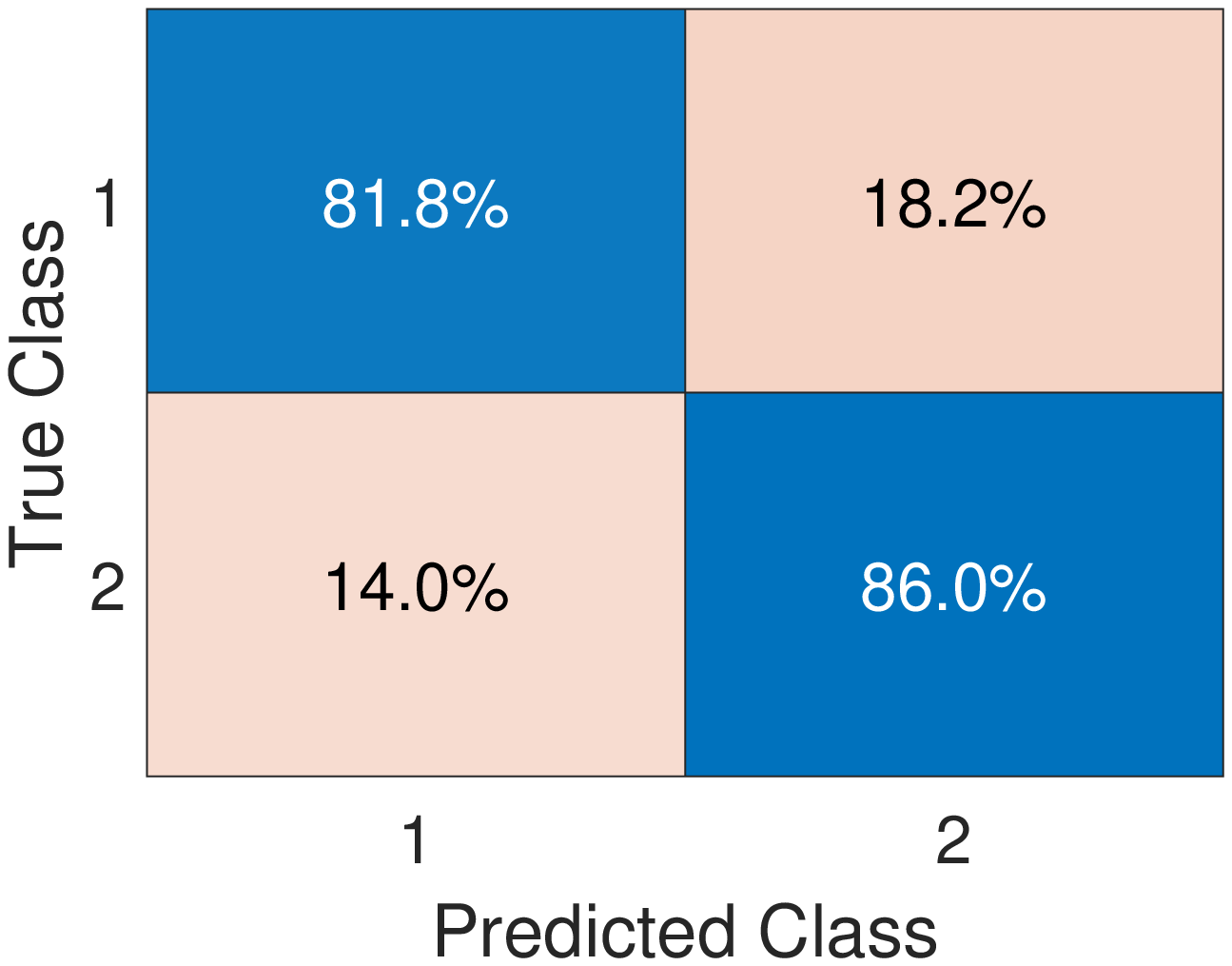} 
		    \caption{HGM confusion matrix}
    \end{subfigure}%
		~
		\begin{subfigure}[t]{0.45\textwidth}
        \centering
         \includegraphics[width=4cm]{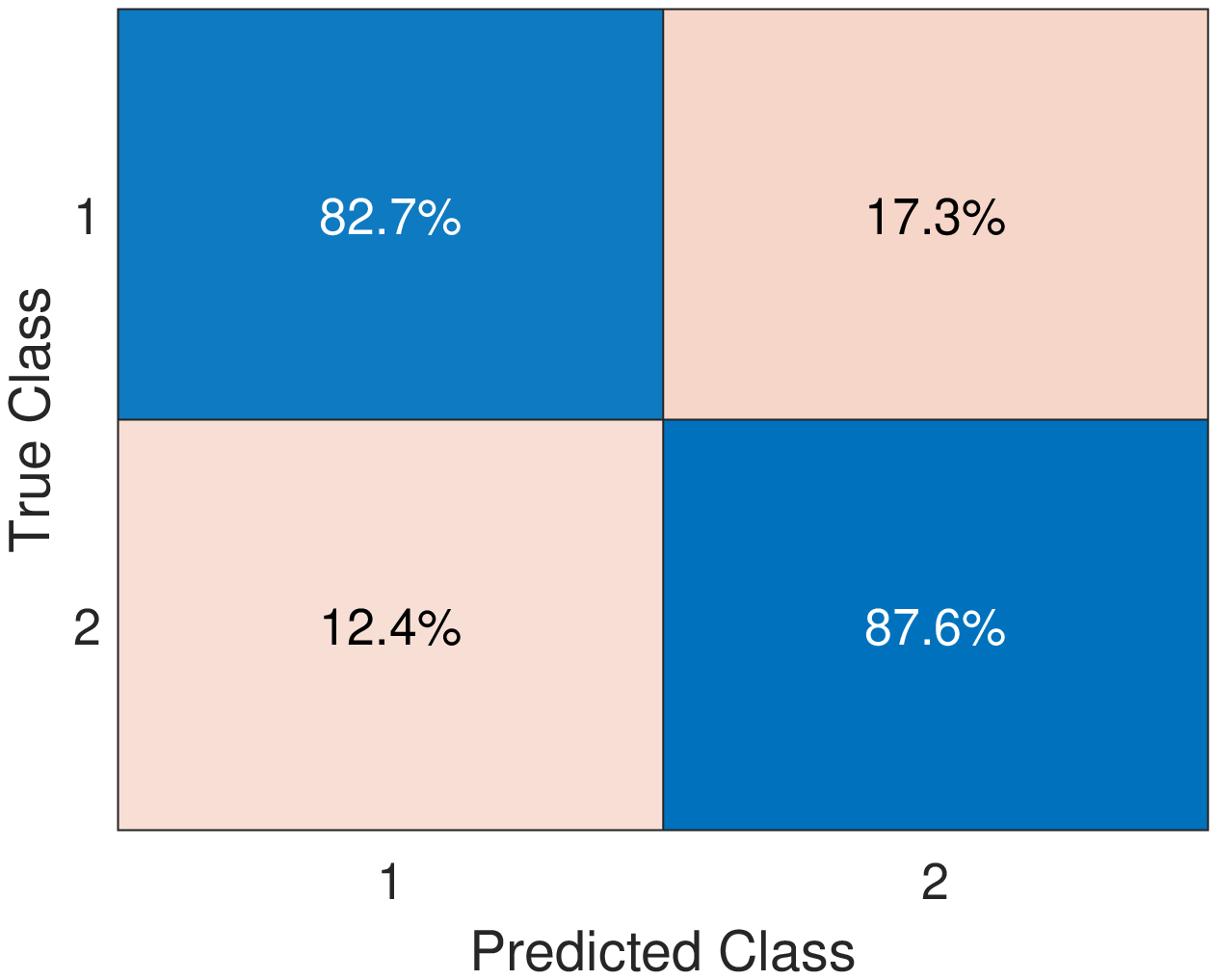} 
		    \caption{HGF confusion matrix}
    \end{subfigure}%
		
		\begin{subfigure}[t]{0.45\textwidth}
        \centering
         \includegraphics[width=4cm]{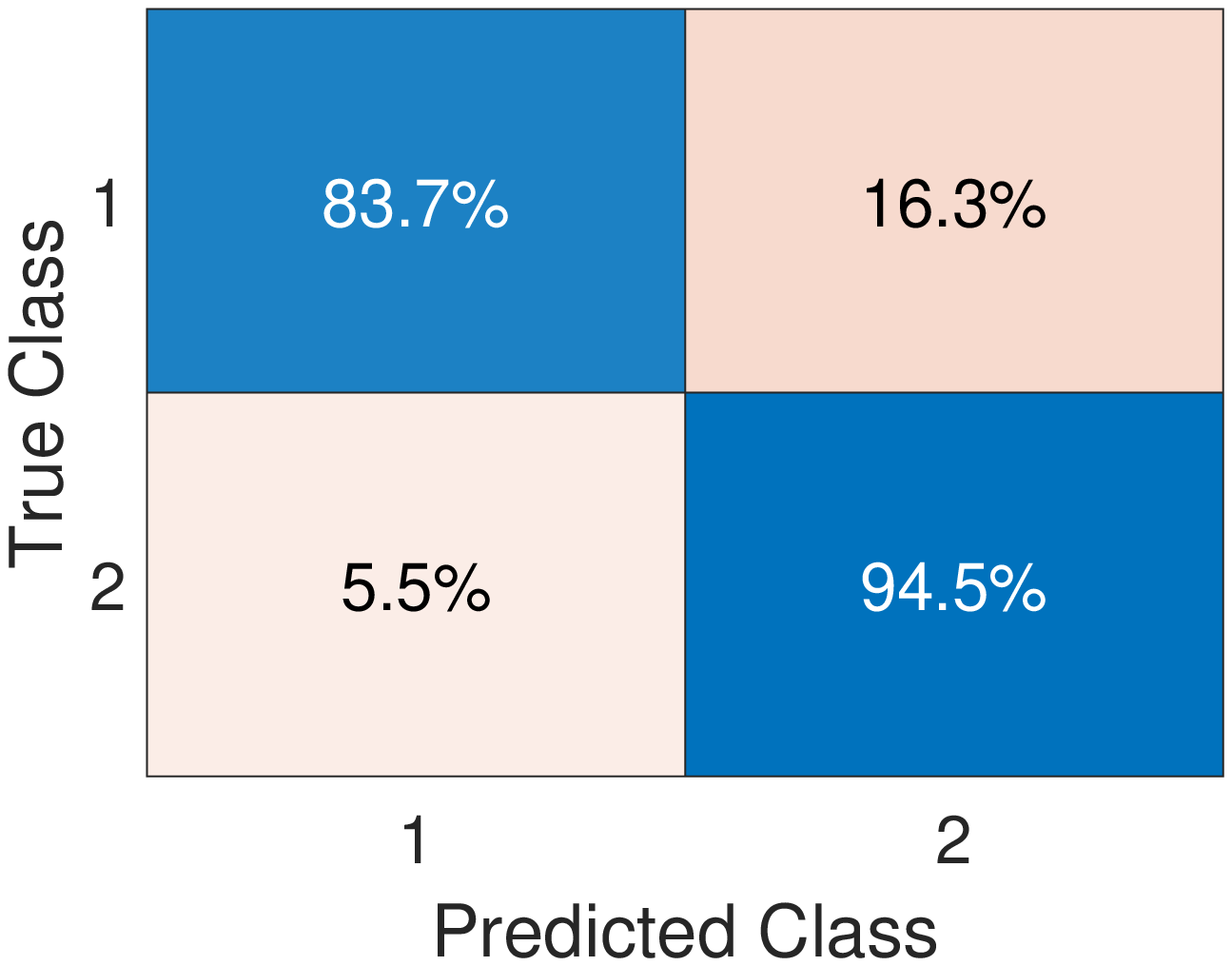} 
		     \caption{SHF confusion matrix}
    \end{subfigure}%
		~
		\begin{subfigure}[t]{0.45\textwidth}
        \centering
         \includegraphics[width=4cm]{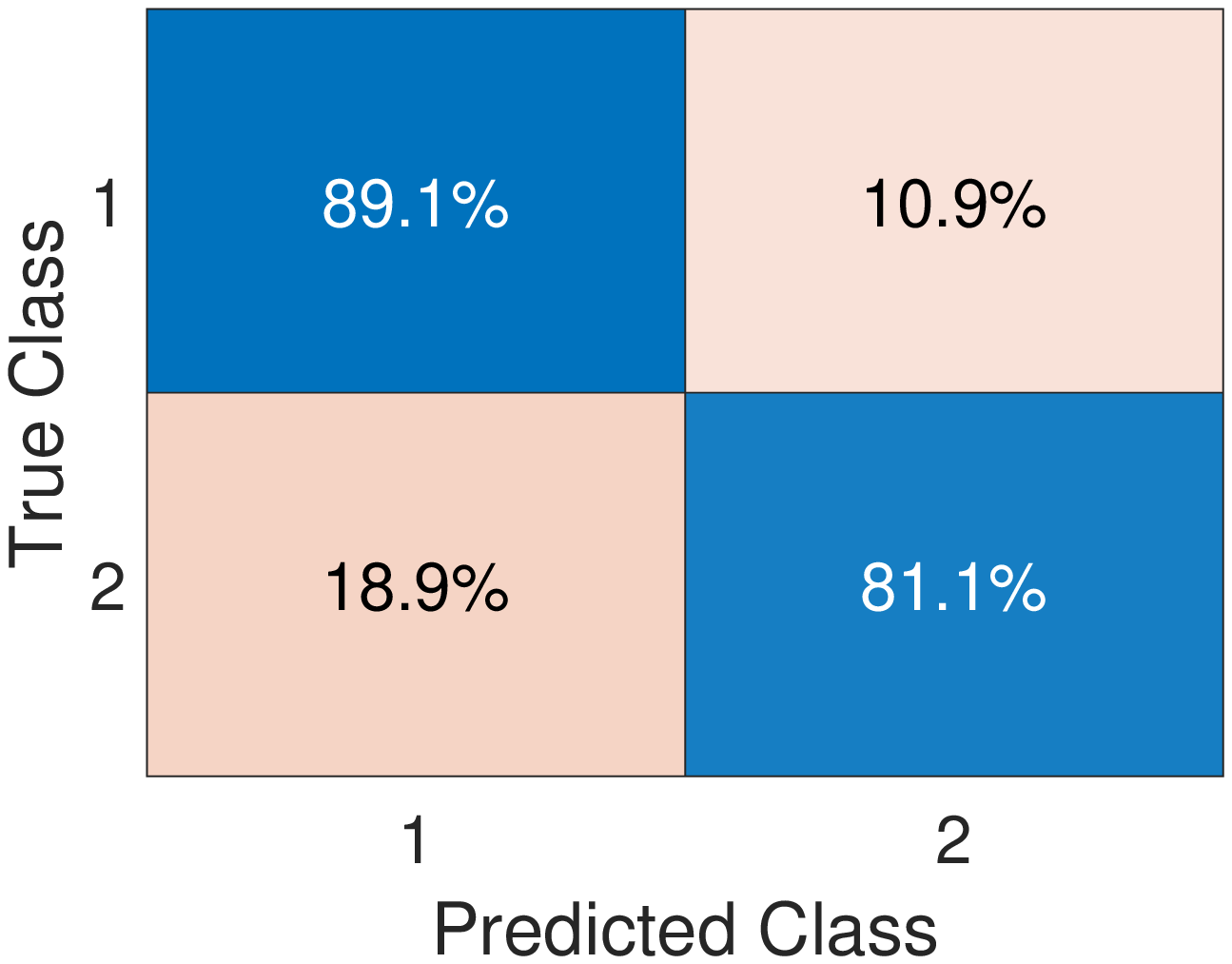} 
		     \caption{GMO confusion matrix}
    \end{subfigure}
		
    \caption{Classification confusion matrices to inform us about the accuracy with which one can identify the existence of different material-types: 1 - Material does not exist, 2 - Material exist}
		\label{Fig015}
\end{figure*}

Eight material-types were randomly selected from all existing material types, in no specific order, to demonstrate the feasibility of this concept. The results for this section are shown in \figurename{ \ref{Fig015}} for the following material types: i) shale (SHL), ii) banded iron formation (BIF), iii) powdery banded iron formation (BPO), iv) goethite ochreous (GOL), v) hematite goethite medium (HGM), vi) hematite goethite friable (HGF), vii) shale ferruginous (SHF), and viii) goethite martite vitreous goethite matrix (GMO). Given the available number of samples, these results demonstrate that MWD can be used to predict the presence/absence of different material-types with accuracies in the order of 80--90\% across the selected material types.

\subsection{Feature Importance and Selection}
The importance of the various features used for assay prediction is compared in this section, to provide insights into the different features' relevance for this application. This is aided by algorithms such as RF providing a measure of predictor importance for process discovery analysis.


In this case, the analysis showed that the top ten features that were highly ranked by RF for Fe predictions in \tindog{} included i) the maximum values of torque, rotationPressure, and pressure ratio, ii) the waveform length of rotationPressure and pressure ratio, iii) the standard deviation of pressure ratio, iv) Hjorth activity of pressure ratio, torque, and sed, and v) median of airPressure. For predicting Fe in \homestead{}, the following features were ranked as top ten by the utilized RF model, including i) the flatness of sed and rop, ii) Hjorth activity of sed, iii) Hjorth mobility and complexity of rop, iv) standard deviation of sed and rop, v) median of airPressure and rotationPressure, and vi) kurtosis of rop. For Sulfur prediction across the combined data from the two regions, the top ten features included all of i) the median of rop, airPressure, torque and sed, ii) Hjorth activity of airPressure, rop, torque and sed, and iii) geometric mean of rop, airPressure. On the other hand, for predicting phosphorus based on the combined data from two regions, the utilized RF model ranked the top features to include i) geometric mean of rop and airPressure, ii) median of rop airPressure and rotationRPM, iii) Hjorth activity of airPressure, rop, and sed, iv) Hjorth complexity of sed, and v) the integral sum of rotationRPM. 

As can be seen from the above list of features, the ranking and the type of top ten contributing features to each assay estimation accuracy would be different from one chemical assay to another and could be even different from one region to another (depending on the distribution of the data and the amount of available data). In the context of the available data within this study, Hjorth parameters (specifically the activity), waveform length, flatness and basic statistics were almost always among the top-ranked features. This does not exclude the importance of the remaining features but is simply intended to highlight the common key features. As the primary goal of this paper was to prove the feasibility of predicting assays and materials from MWD data, a more thorough analysis of the feature importance and ranking will be included in future work on a larger dataset.

\section{Discussion}
The results presented in this paper indicate that the approach of mapping MWD to materials types and chemical assay values is feasible, leading to the conclusion that machine learning algorithms can be employed to predict materials and assays. While this was only demonstrated on eight materials types, the findings do generalize across a broader set of materials (excluded for brevity). It is important to mention here that our findings also suggest that the accuracy of this mapping process is primarily dependent on (i) assay/material type and its distribution across the mining sites from which the data was acquired and (ii) the amount of available data. It was also found that the choice of the regression model, among GP, SVM and RF models, did not significantly alter the findings, as the quality of the estimation performed by these machine learning models mainly depended on the quality of the extracted features. 

The analysis started with Fe estimation using GP, SVM, and RF models, with results suggesting that Fe estimation across \tindog{} was more accurate than that across \homestead{}. The distribution of Fe across the two mining sites was then examined in \figurename{ \ref{Fig006}} and it was found that the two mining sites had significantly different distributions of iron as the histogram of Fe across \homestead{} spiked much higher than that across \tindog{} (larger kurtosis value). This in turn explains part of the variability of the results from both sites. Further variability appears to come from limitations on the available data, with approximately 2000 samples available for each site resulting in fewer samples within some classes (waste, low-grade, high-grade). It is important to note that the largest Fe prediction errors were actually made within the waste class, as was shown in \figurename{ \ref{Fig008}}, while much more consistent predictions were made along low-grade and high-grade iron where the predicted values would actually be relevant. 

In terms of Phosphorus (P), the largest prediction errors were made along values larger than 0.35, as when combining data from \tindogandhomestead{} it was found that out of the set of nearly 4000 samples, only around 11 Phosphorus samples were available with values larger than 0.35. Hence, there was not enough data for the GP, SVM and RF models to learn the relationship between MWD and high Phosphorus values. However, the overall prediction accuracy of Phosphorus suggests its estimation is useful, with a Pearson correlation coefficient of 0.81, RMSE of 0.3, and a $p$-value of $<$ 0.001. This indicates significant correlations between laboratory measurements and estimations by our models. In terms of Sulfur (S), the predictions shown in this paper were found to be most accurate when combining the data from the two mining sites, with the best results achieved by RF with a correlation coefficient of 0.92, RMSE of 0.002604, and a $p$-value of $<$ 0.001. 


The effectiveness of using knowledge about one assay type to assist the prediction of other assays was investigated. The hypothesis, in this case, is that if one algorithm or sensor can predict one assay type accurately, then one can make use of this knowledge to better predict other assays. For this part of the analysis, when using the MWD features with an RF model, the combined \tindogandhomestead{} data had a correlation coefficient of 0.65 with an RMSE of 1.6 when predicting Al\textsubscript{2}O\textsubscript{3}. However, these results were significantly enhanced by adding either Fe ($r=0.9$, $\text{RMSE}=0.92$) or SiO\textsubscript{2} ($r=0.85$, $\text{RMSE}=1.1$), as was shown in \figurename{ \ref{Fig014}}. The reason for selecting these two assays to augment MWD data when predicting Al\textsubscript{2}O\textsubscript{3} is that high correlations were observed between these assays, as depicted in \figurename{ \ref{Fig013}}. Hence, the knowledge of either of these assays supported the model to more accurately estimate the Al\textsubscript{2}O\textsubscript{3} values.

In the final part of the experiments, we presented a proof of concept that MWD can also predict the presence or absence of material types. Eight different materials types including i) SHL, ii) BIF, iii) BPO, iv) GOL, v) HGM, vi) HGF, vii) SHF and viii) GMO were randomly selected to demonstrate this. The confusion matrices generated when using the SVM as a classifier to predict material presence or absence showed, in general, accuracies over 80\% achieved with the available data.

\section{Conclusion}
In this paper, a proof of concept has been presented for using machine learning and MWD data to predict the presence or absence of material types and estimate chemical assay values. It is important to note that while different applications of MWD have been previously considered in the literature, the study presented in this paper is the first to use MWD for material logging and assaying purposes, to the best of the authors' knowledge. The findings of this study strongly support the feasibility of the proposed approach, with results showing correlations between MWD features and assays types of up to 0.92 for individual assays. The analysis has also shown that the accuracy depends on the distribution of the data and the assay type being predicted. It was also demonstrated that the presence or absence of material types can be predicted while drilling by using MWD data based on a limited dataset.

The findings of this paper are important to the mining industry as the timeliness and quality of these estimations cascade through the downstream mining processes. Knowledge of material-types and chemical assays can play a significant role in mining, guiding the drilling process, orebody modeling, and providing chemistry data with down-hole resolution. The generated predictions and estimates of material types and assays can also further help guide mine planning. The work in this field continues as the authors plan to investigate the impact of using deep learning models on much larger datasets from a wider range of sites. Further studies will be conducted on automatic feature extraction in comparison to the handcrafted feature extraction approach utilized in this paper, which demonstrated the feasibility of using this MWD data for logging and assaying.

\begin{acknowledgements}
This work has been supported by the Australian Centre for Field Robotics and the Rio Tinto Centre for Mine Automation. The authors would also like to acknowledge the support of Anna Chlingaryan and Katherine Silversides in the manuscript review and editing process.

\end{acknowledgements}

%
%


\bibliographystyle{MG}       
{\footnotesize
\bibliography{MG_template}}   

%

\end{document}